\documentclass{article}
\usepackage{iclr2027_conference,times}

\usepackage{amsmath,amsfonts,bm}

\def\eqref#1{equation~\ref{#1}}

\def\plaineqref#1{\ref{#1}}

\def\1{\bm{1}}

\DeclareMathAlphabet{\mathsfit}{\encodingdefault}{\sfdefault}{m}{sl}
\SetMathAlphabet{\mathsfit}{bold}{\encodingdefault}{\sfdefault}{bx}{n}

\usepackage{hyperref}
\usepackage{url}
\usepackage{booktabs}
\usepackage{tabularx}
\usepackage{colortbl}
\usepackage{arydshln} %
\usepackage{xcolor}

\usepackage{enumitem}
\usepackage{amsmath,amssymb}
\usepackage{graphicx}
\usepackage{tikz}
\usetikzlibrary{arrows.meta,positioning}

\title{EviSplat: Preserving Multi-View Evidence in 3D Gaussian Splatting for Open-Vocabulary Segmentation}

\author{\normalfont
\begin{minipage}{0.97\textwidth}
\centering
Sungho Moon$^{1}$ \quad Kota Shimomura$^{2,5}$ \quad Junwoo Park$^{1}$ \quad Wonhyeok Choi$^{1}$\\[3pt]
Seunghun Lee$^{3}$ \quad Takayoshi Yamashita$^{2}$ \quad Sunghoon Im$^{4}$\\[7pt]
{\small $^{1}$DGIST, Daegu, Republic of Korea\\
$^{2}$Chubu University, Japan \qquad $^{3}$HUAWEI, Singapore\\
$^{4}$KAIST, Daejeon, Republic of Korea \qquad $^{5}$Elith Inc., Japan}
\end{minipage}}

\iclrfinalcopy
\hypersetup{
  hidelinks,
  pdftitle={EviSplat: Preserving Multi-View Evidence in 3D Gaussian Splatting for Open-Vocabulary Segmentation},
  pdfauthor={Sungho Moon, Kota Shimomura, Junwoo Park, Wonhyeok Choi, Seunghun Lee, Takayoshi Yamashita, Sunghoon Im}
}

\begin{document}

\maketitle
\fancyhead{}
\renewcommand{\headrulewidth}{0pt}

\begin{abstract}
Open-vocabulary 3D scene understanding enables object localization and segmentation from free-form text queries without a fixed category vocabulary. 
Many recent methods build on 3D Gaussian Splatting and consolidate multi-view observations, such as masked crops from individual views, into language features or compact object descriptors before the query is known. 
However, observations of the same object vary across viewpoints and are not equally informative: some reveal cues relevant to a particular query, whereas others provide incomplete or misleading evidence. 
Pre-query consolidation can therefore suppress cues on which a later query depends. 
We introduce \emph{EviSplat}, which preserves individual observation features as evidence for later text queries. 
EviSplat retains individual observation features within class-agnostic 3D instances that represent objects, object parts, or background regions. 
It also learns, for each Gaussian, a distribution describing which visual appearances its observations support. 
Given a text query, EviSplat scores each instance using its most relevant observations. 
It then computes a score for each Gaussian by combining instance-level relevance with locally supported evidence, weighted by how often and how unambiguously that Gaussian was observed. 
Different queries can thus draw on different visual cues from the same preserved evidence. 
Experiments across diverse datasets and evaluation protocols demonstrate state-of-the-art performance, supporting the benefit of preserving multi-view evidence until query time and aggregating it according to the query.
\end{abstract}

\section{Introduction}
\label{sec:intro}

Open-vocabulary 3D scene understanding aims to make a reconstructed scene queryable with free-form text rather than a fixed set of category labels~\citep{kerr2023lerf,qin2024langsplat,shi2024legaussians,wu2024opengaussian}.
We focus on open-vocabulary 3D object segmentation, where the goal is to localize and segment the object referred to by a text query~\citep{qin2024langsplat,wu2024opengaussian,cen2025laga}.
Each masked crop from a source view forms an observation, which we encode as a CLIP feature. %
These observations can reveal different appearance cues as the viewpoint~\citep{cen2025laga} and mask granularity change~\citep{peng2024gags}.%
Other observations are incomplete or misleading because of occlusion, blur, or inaccurate masks~\citep{huang2025openinsgaussian,cheng2024occamlgs,wang2025vala}. %
The view-specific cues carried by these observations constitute the multi-view evidence available to a later text query.

\begin{figure}[t]
\centering
\includegraphics[width=\linewidth]{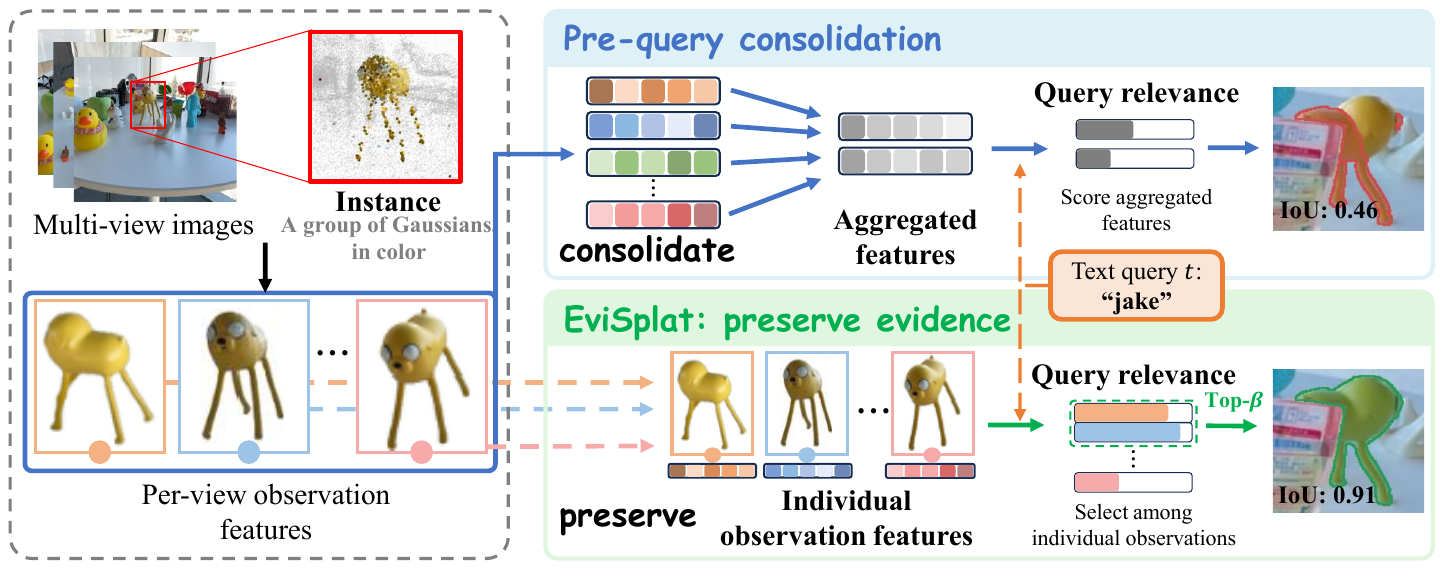} %
\caption{\textbf{Pre-query consolidation vs.\ EviSplat: preserve evidence.}
Top: features aggregated across observations are scored against the query.
Bottom: EviSplat retains individual observation features for query-dependent selection and score aggregation.
Crops illustrate the source observations; feature vectors and relevance bars are schematic.
For the query ``jake'' in the same view, the masks show LaGa (top, IoU $0.46$) and the full EviSplat model (bottom, IoU $0.91$).}
\label{fig:motivation}
\end{figure}

Recent methods represent scenes with 3D Gaussian Splatting (3DGS)~\citep{kerbl20233d} and store language cues either as features learned at individual Gaussians~\citep{qin2024langsplat,shi2024legaussians,li2026langsplatv2} or as descriptors attached to class-agnostic 3D instances~\citep{wu2024opengaussian,cen2025laga}, which represent objects, object parts, or background regions as groups of Gaussians.
A common design combines features from multiple observations before the query is known, through feature distillation or descriptor clustering.
However, a feature that represents an object well overall may not preserve the cues needed to distinguish it for a particular query.
Fig.~\ref{fig:motivation} (top) illustrates scoring these aggregated features against the query.
Once individual observation features are replaced by these summaries, the query cannot select among the original observations, potentially weakening query-relevant cues.

To address this limitation, we propose \emph{EviSplat}, which keeps individual observation features available for query-dependent selection.
Using LaGa's class-agnostic grouping~\citep{cen2025laga}, EviSplat organizes these features into instance observation banks (Fig.~\ref{fig:motivation}, bottom).
Since different observations may cover different subsets of an instance's Gaussians, an instance-level score alone cannot distinguish where query-relevant cues are supported. 
EviSplat therefore learns an appearance distribution for each Gaussian to capture its locally supported evidence.
Given a query, query-guided evidence aggregation forms an instance-level score from the most relevant observations and refines each Gaussian's score with local evidence weighted by observation reliability.

\begin{samepage}
We evaluate EviSplat on LERF-OVS, 3D-OVS, and ScanNet for open-vocabulary 3D segmentation.
On LERF-OVS, EviSplat improves mean IoU from $63.6$ to $70.7$, an $11.2\%$ relative gain over our LaGa baseline~\citep{cen2025laga}.
Our contributions are threefold:
\begin{enumerate}[leftmargin=1.6em,itemsep=1pt,topsep=2pt]
\item We propose a two-level evidence representation that combines \emph{instance observation banks} with a \emph{Gaussian evidence field}, encoding which visual appearances are supported at each Gaussian.
\item We propose \emph{query-guided evidence aggregation}, which selects the most relevant observations of each instance and adds reliable Gaussian-local evidence on top of the resulting instance score. %
\item We demonstrate that \emph{EviSplat} improves open-vocabulary 3D segmentation across multiple benchmarks under both rendered-view and point-level evaluations, and that instance-level observation selection and Gaussian-local evidence provide complementary gains.
\end{enumerate}
\end{samepage}

\section{Related Work}
\label{sec:related}

\paragraph{Language fields and instance grouping.}
Language fields incorporate pretrained image--language features into 3D scene representations that can be queried using natural language.
LERF \citep{kerr2023lerf} distills multi-scale CLIP features into a NeRF~\citep{mildenhall2021nerf}, and LangSplat \citep{qin2024langsplat} extends this approach to 3DGS with hierarchical mask supervision and a scene-specific feature autoencoder.
LEGaussians \citep{shi2024legaussians} compresses CLIP and DINO features through quantization, while LangSplatV2 \citep{li2026langsplatv2} represents language features using sparse coefficients over a shared dictionary.
VLGS \citep{peng2025vlgs} improves semantic rendering through cross-modal rasterization and camera-view blending.

Class-agnostic representations group scene elements into instances without assigning semantic categories, allowing language features to be associated with these instances. %
Gaussian Grouping \citep{ye2024gaussiangrouping} learns identity encodings from SAM masks associated across views by a zero-shot tracker, and SAGA \citep{cen2025saga} learns scale-gated affinity features.
OpenGaussian \citep{wu2024opengaussian} connects instance features to language through codebook discretization and 3D--2D association.
GALA \citep{alegret2026gala} couples instance and language codebooks, while OpenGaFF \citep{li2026opengaff} retrieves language features from a structured codebook using a geometry-conditioned feature field.

\paragraph{Multi-view association and semantic aggregation.}
Across views, observations of the same object differ in visibility, mask extent, and semantic features.
Multi-view association links corresponding observations, while semantic aggregation combines their information into a 3D representation.
GAGS \citep{peng2024gags} adapts supervision granularity, and CCL-LGS \citep{tian2025cclgs} associates SAM masks across views with a zero-shot tracker before contrastive codebook learning.
OpenInsGaussian \citep{huang2025openinsgaussian} combines context-aware feature extraction with attention-driven fusion, while ReLaGS \citep{xie2026relags} aggregates language features within a hierarchical Gaussian scene.
Direct feature registration offers an alternative to rendering-based feature distillation: Occam's LGS \citep{cheng2024occamlgs} uses rendering-weighted aggregation, Dr.~Splat \citep{kim2025drsplat} assigns embeddings to dominant ray-intersected Gaussians, and VALA \citep{wang2025vala} combines visibility gating with robust aggregation.
At the object level, LaGa \citep{cen2025laga} clusters and reweights multiple semantic descriptors.
Arafa and Stricker~\citep{arafa2025beyond} retain per-view CLIP embeddings for each object and average the top-$k$ query-relevance scores to obtain an object-level retrieval score.

\section{Preliminaries}
\label{sec:method_prelim}

\paragraph{3D Gaussian Splatting.}
3D Gaussian Splatting (3DGS) \citep{kerbl20233d} represents a scene explicitly as a set $\mathcal G$ of anisotropic 3D Gaussians, each parameterized by a center, a covariance matrix, an opacity, and a view-dependent color given by spherical-harmonic coefficients.
To render a pixel $p$ in view $v$, the Gaussians overlapping its ray are sorted front to back and $\alpha$-blended.
The contribution of Gaussian $g\in\mathcal G$ is weighted by
\begin{equation}
A_g^{(v)}(p)=\alpha_g^{(v)}(p)\prod_{g'\prec_v g}\bigl(1-\alpha_{g'}^{(v)}(p)\bigr), %
\label{eq:compositing}
\end{equation}
where $\alpha_g^{(v)}(p)$ is the projected opacity of $g$ at pixel $p$ in view $v$, and $g'\prec_v g$ denotes a Gaussian lying in front of $g$ along the ray. %

\paragraph{Language features and query relevance.}
Language-embedded 3DGS methods~\citep{qin2024langsplat,cen2025laga} typically localize a text query in three stages.
First, each source view is segmented by SAM \citep{kirillov2023segment} at a set $\mathcal S$ of semantic scales corresponding to different mask granularities~\citep{qin2024langsplat}.
Each masked crop is then encoded as a unit-norm feature by a frozen CLIP image encoder~\citep{radford2021learning}.
Second, before a query is known, the features collected across views are consolidated into aggregated features attached to the 3D scene, such as one language feature per Gaussian~\citep{qin2024langsplat} or a set of descriptors per 3D object~\citep{cen2025laga}.
Third, given a text query $t$, each aggregated feature $\mathbf x$ is scored using the LERF relevance function~\citep{kerr2023lerf},
\begin{equation}
r(\mathbf x,t)
=
\min_{t'\in\mathcal T_{\mathrm{canon}}}
\frac{\exp(\mathbf x^\top\boldsymbol\phi_t)}
     {\exp(\mathbf x^\top\boldsymbol\phi_t)+\exp(\mathbf x^\top\boldsymbol\phi_{t'})}
\in(0,1),
\label{eq:relevance}
\end{equation}
where $\boldsymbol\phi_t$ is the unit-norm CLIP text embedding of $t$, and $\mathcal T_{\mathrm{canon}}$ is a fixed set of canonical negative phrases such as ``object''.
In particular, $r(\mathbf x,t)>0.5$ if and only if $\mathbf x$ is more similar to the query than to every canonical phrase.
Features with high relevance are subsequently selected to localize the query in the 3D scene.

\section{Method}
\label{sec:method}
\begin{figure}[t]
\centering
\includegraphics[width=\linewidth]{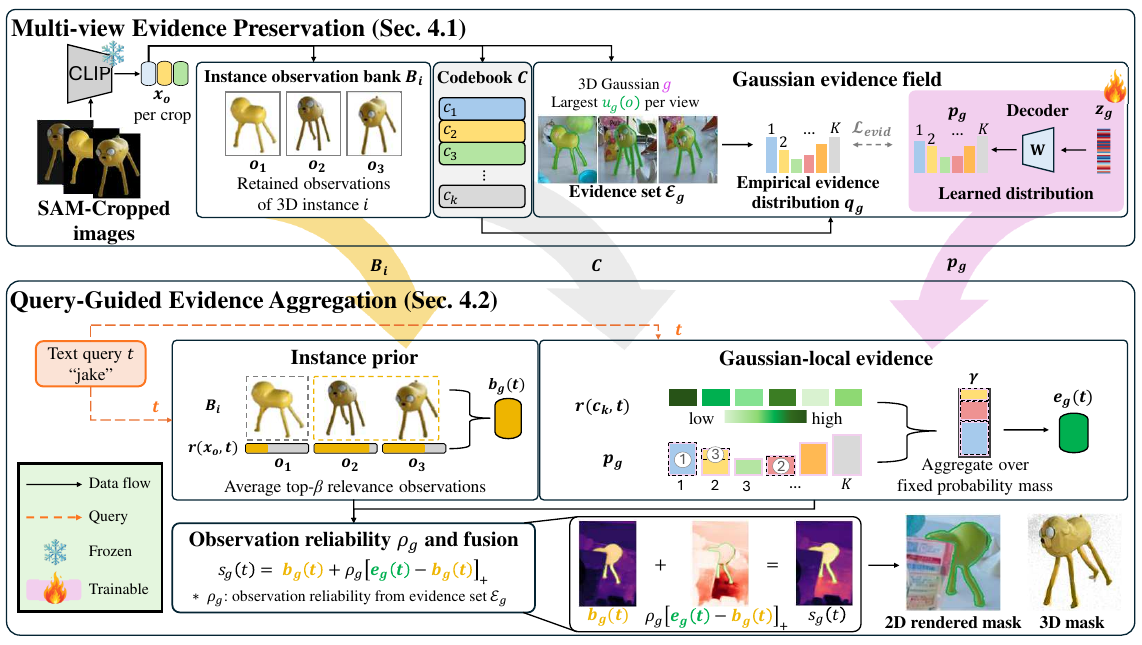} %
\caption{\textbf{Overview of EviSplat.}
Top: multi-view evidence preservation, illustrated for one instance $i$ (three of its retained observations shown) and one Gaussian $g$ seen in three views.
Bottom: query-guided evidence aggregation for the query ``jake''.
Histograms show a subset of the $K$ prototype bins, colored consistently with the codebook; only the latents and decoders (purple) are trained.
The two stages share the observation banks $\mathcal B_i$, codebook $\mathcal C$, and learned distributions $p_g$.}
\label{fig:main_pipeline}
\end{figure}

EviSplat preserves individual observation features and aggregates their relevance at query time, assigning each Gaussian an evidence-aggregated score $s_g(t)$ in place of the relevance of aggregated features $r(\mathbf x,t)$ for final localization.
As illustrated in Fig.~\ref{fig:main_pipeline}, EviSplat comprises two stages.
\emph{Multi-view evidence preservation} (\S\ref{sec:method_evidence_field}) constructs two representations before querying. %
The instance observation banks retain individual multi-view features for query-dependent selection, while the
Gaussian evidence field records a distribution of the appearances supported at each Gaussian.%
After this stage, \emph{Query-guided evidence aggregation} (\S\ref{sec:method_query_scoring}) scores both representations against an incoming query and computes the final scores of individual Gaussians.

\subsection{Multi-view Evidence Preservation} %
\label{sec:method_evidence_field}
We construct the instance observation banks and Gaussian evidence field from the per-view SAM masks and CLIP features described in \S\ref{sec:method_prelim}. %
Unless stated otherwise, all quantities are scale-specific; only the
Gaussian latent $\mathbf z_g$ is shared across semantic scales.

\paragraph{Instance observation banks.}
We collect observations across source views into a set $\mathcal O$. Each observation $o\in\mathcal O$ is associated with its source view and consists of a SAM mask $m_o$ and the unit-norm CLIP feature $\mathbf x_o$ of its masked crop. %
We group observations of the same scene element without merging their
features, allowing a later query to select among the retained evidence. %
To associate masks across views, we adopt the class-agnostic affinity
field of LaGa~\citep{cen2025laga}, which assigns each Gaussian $g$ an
affinity feature $\mathbf a_g$ such that Gaussians covered by the same
mask have similar features.
We render this field and pool it over each mask $m_o$ to obtain an
observation-level affinity feature $\mathbf a_o$. %
HDBSCAN~\citep{mcinnes2017hdbscan} then clusters
$\{\mathbf a_o\}_{o\in\mathcal O}$ across views.
Each resulting cluster defines a class-agnostic \emph{3D instance} $i\in\mathcal I$, and its members form the \emph{instance observation bank} $\mathcal B_i\subseteq\mathcal O$. Depending on the semantic scale, an instance may represent an object,
an object part, or a background region; we do not assume that every instance corresponds to a complete object.

\paragraph{Evidence sets.}
The observation banks preserve evidence at the instance level but do not specify where each observation is supported within an instance.
We therefore associate each Gaussian $g$ with at most one observation from each source view.
Let $\mathcal V$ denote the set of source views. For each $v\in\mathcal V$, let $\mathcal O_v\subseteq\mathcal O$ be the observations extracted from $v$. %
For an observation $o\in\mathcal O_v$, we define its association with $g$ as the fraction of the rendered mass of $g$ that falls within $m_o$:
\begin{equation}
u_g(o)=\frac{\sum_{p \in m_o}A_g^{(v)}(p)}{\sum_{p \in \mathcal{P}_v}A_g^{(v)}(p)}, %
\label{eq:ownership}
\end{equation}
\noindent where $A_g^{(v)}(\cdot)$ is the compositing weight of Eq.~\ref{eq:compositing}, $\mathcal P_v$ is the set of all pixels of view $v$, and we set $u_g(o)=0$ when $g$ contributes to no pixel in that view.
Unlike a test on the projected Gaussian center, this association accounts for occlusion and transmittance through the compositing weights.

For each view with $\mathcal O_v\ne\varnothing$, we select one observation $o_g^{(v)}$ with the largest association, breaking ties consistently. The evidence set retains selections with positive association: %
\begin{equation}
  \begin{aligned}
  o_g^{(v)}
  &\in \operatorname*{arg\,max}_{o\in\mathcal O_v}u_g(o), \qquad \mathcal E_g
  &=\{o_g^{(v)}\mid v\in\mathcal V,\
  \mathcal O_v\ne\varnothing,\ u_g(o_g^{(v)})>0\}.
  \end{aligned}
\label{eq:evidence_set}
\end{equation}
For a selected observation $o\in\mathcal E_g\cap\mathcal O_v$, we confidence in this Gaussian-mask association by its relative margin over the runner-up in the same view, %
\begin{equation}
\omega_g(o)=1-\frac{\max_{o'\in\mathcal O_v\setminus\{o\}}u_g(o')}{u_g(o)}\;\in[0,1],
\label{eq:margin}
\end{equation}
If $o$ is the only mask in its view, we set $\omega_g(o)=1$. We obtain $w_g(o)$ by normalizing the assignment margins across Gaussians with selected observations in the same view. When their sum is positive, this gives each view unit total weight, independently of how many Gaussians receive observations. %

\paragraph{Evidence distribution.}
A single mean feature does not uniquely determine the underlying observation distribution. 
Let $P_g$ denote the empirical distribution of features in $\mathcal E_g$, with probabilities proportional to $w_g(o)$, and let $\bar{\mathbf x}_g$ be its normalized mean.
Because query relevance is nonlinear, scoring this mean generally differs from averaging the individual relevance scores: %
\begin{equation}
\bar{\mathbf x}_g
=
\frac{\mathbb E_{\mathbf x\sim P_g}[\mathbf x]}
     {\left\|\mathbb E_{\mathbf x\sim P_g}[\mathbf x]\right\|_2},
\qquad
r(\bar{\mathbf x}_g,t)
\neq
\mathbb E_{\mathbf x\sim P_g}[r(\mathbf x,t)].
\label{eq:noncommutativity}
\end{equation}
Different observation distributions can share the same normalized mean while yielding different relevance distributions for a query. To retain information for query-dependent selection, we therefore approximate the observation distribution with a weighted histogram over visual prototypes. %
Specifically, we cluster $\{\mathbf x_o\}_{o\in\mathcal O}$ using spherical $K$-means to construct a frozen codebook $\mathcal C=\{\mathbf c_k\}_{k=1}^{K}$ of unit-norm visual prototypes shared by all Gaussians in the scene. These prototypes represent observed appearances without relying on a predefined category vocabulary.
Let $\kappa(o)=\arg\max_{k}\mathbf c_k^\top\mathbf x_o$ be the prototype nearest to observation $o$.
For evidence sets with positive total weight, the empirical distribution $q_g$ is the normalized weighted histogram over prototypes:%
\begin{equation}
q_g(k)=\frac{\sum_{o\in\mathcal E_g, \kappa(o)=k}w_g(o)}{\sum_{o\in\mathcal E_g}w_g(o)},
\label{eq:exact_target}
\end{equation}
Each selected observation contributes a weighted vote to its nearest prototype, and $q_g$ records the normalized distribution of these votes. %

Storing $q_g$ sparsely requires more entries as observations support additional prototypes. To control this storage, we learn a latent vector $\mathbf z_g\in\mathbb R^D$ for each Gaussian, shared across scales, and a linear decoder $\mathbf W\in\mathbb R^{K\times D}$ for each scale to approximate $q_g$: %
\begin{equation}
p_g=\operatorname{softmax}\!\left(\frac{\mathbf W\mathbf z_g}{\lVert\mathbf z_g\rVert_2}\right),
\qquad
\mathcal L_{\mathrm{evid}}
=
-\,\mathbb E_{g}\Bigl[\,\sum_{k=1}^{K}q_g(k)\log p_g(k)\Bigr],
\label{eq:evidence_loss}
\end{equation}
where Gaussians are sampled with probability proportional to $\sum_{o\in\mathcal E_g}w_g(o)$, so that every observation contributes to the loss in proportion to its weight. 
We call the map $g\mapsto p_g$ the \emph{Gaussian evidence field}.
The latents and decoders are the only trained parameters of EviSplat; the 3DGS scene, the CLIP encoder, and the affinity field remain frozen, and training uses neither text queries nor a label vocabulary.

\subsection{Query-Guided Evidence Aggregation}
\label{sec:method_query_scoring}
Given a text query $t$, every Gaussian $g$ receives the score
\begin{equation}
s_g(t)=b_g(t)+\rho_g\,\max\bigl(e_g(t)-b_g(t),\,0\bigr).
\label{eq:fusion}
\end{equation}
The \emph{instance prior} $b_g(t)$ provides a base score for $g$ by combining instance relevance scores according to its affinity-based assignments. %
The \emph{Gaussian-local evidence} $e_g(t)$ is the relevance of the appearances observed at $g$ itself, computed from the evidence field.
The \emph{observation reliability} $\rho_g\in[0,1]$, computed from the evidence sets, determines how much the local evidence may raise the prior.
We define the three terms in turn.

\paragraph{Instance prior $b_g(t)$.}
We apply the relevance of Eq.~\ref{eq:relevance} to the individual features of each bank rather than to a consolidated descriptor. %
Average pooling over a bank can dilute useful cues, whereas max pooling is sensitive to individual noisy observations, as also reported by LaGa~\citep{cen2025laga}. %
For a set of $n$ scores, let $\operatorname{mean}_\beta$ denote the mean of its $\lceil\beta n\rceil$ largest elements, with $\beta\in(0,1]$; $\beta=1$ gives average pooling and $\beta\to0$ gives max pooling.
To pass instance scores to Gaussians, we compute query-independent affinity assignments once,
\begin{equation}
\pi_g(i)\propto\exp\bigl(\cos(\mathbf a_g,\mathbf a_i)/T\bigr),\qquad\sum_{i\in\mathcal I}\pi_g(i)=1.
\label{eq:assignment}
\end{equation}
This averaging is used only for affinity-based assignment; the CLIP features in each bank remain separate.
Here $\mathbf a_i$ is the mean of $\mathbf a_o$ over $\mathcal B_i$, $\cos(\cdot,\cdot)$ denotes cosine similarity, and $T>0$ controls assignment sharpness; only the observation scores change with query.
The instance relevance and its assignment-weighted Gaussian prior are
\begin{equation}
R_i(t)=\operatorname{mean}_\beta\{r(\mathbf x_o,t)\mid o\in\mathcal B_i\},\qquad
b_g(t)=\sum_{i\in\mathcal I}\pi_g(i)R_i(t).
\label{eq:instance_prior}
\end{equation}

\paragraph{Gaussian-local evidence $e_g(t)$.}
The instance prior draws on instance-level observations, whereas the evidence field records the appearances supported at Gaussian $g$ itself. %
Here the relevance is applied to the codebook prototypes, $r(\mathbf c_k,t)$, while $p_g$ remains fixed.
Expected relevance averages over the full distribution, while maximum relevance ignores probability support. Our component analysis compares both alternatives with fixed-mass aggregation (Tab.~\ref{tab:lerf_component}).%
We therefore take the largest average relevance attainable with a fixed amount $\gamma\in(0,1]$ of probability mass:
\begin{equation}
e_g(t)=\max_{\delta}\;\frac{1}{\gamma}\sum_{k=1}^{K}\delta(k)\,r(\mathbf c_k,t)
\qquad\text{s.t.}\qquad
0\le\delta(k)\le p_g(k),\quad \sum_{k=1}^{K}\delta(k)=\gamma .
\label{eq:gaussian_local_evidence}
\end{equation}
Here, $\delta(k)$ is the probability mass selected from prototype $k$.
The maximum is attained greedily: visiting the prototypes in order of decreasing relevance (the relevance bar above each bin of $p_g$ in Fig.~\ref{fig:main_pipeline}), each contributes its full probability until the budget $\gamma$ is reached, and the last one contributes only the remainder.
A highly relevant prototype with little mass therefore contributes only in proportion to that mass, and $\gamma=1$ recovers the expected relevance.
Thus, $\beta$ selects a fraction of observations, while $\gamma$ selects a fraction of probability mass.
The number of selected prototypes can vary across Gaussians: a concentrated distribution may fill the probability budget with one prototype, whereas a diffuse distribution may require several.

\paragraph{Observation reliability $\rho_g$ and fusion.}
Two Gaussians can have similar appearance distributions even when one is supported by far fewer observations.
To account explicitly for observation count and assignment ambiguity, we derive a reliability weight from the evidence set:%
\begin{equation}
\rho_g=\frac{n_g}{n_g+\bar n}\;\bar\omega_g\;\in[0,1],
\qquad
n_g=|\mathcal E_g|,
\qquad
\bar\omega_g=\frac{1}{n_g}\sum_{o\in\mathcal E_g}\omega_g(o),
\label{eq:observation_reliability}
\end{equation}
where $\bar n$ is the mean of $n_g$ over the Gaussians with a non-empty evidence set, and $\rho_g=0$ if $\mathcal E_g$ is empty.
The factor $n_g/(n_g+\bar n)$ downweights Gaussians with fewer observations relative to the scene average, while the mean assignment margin $\bar\omega_g$ reduces reliability when associations are ambiguous. We average the margins rather than $w_g(o)$, whose magnitude also depends on the total assignment margin in its source view.
Like the assignment, the reliability does not depend on the query and is computed once; since $\bar n$ comes from the scene itself, it introduces no hyperparameter.

Because local observations may be incomplete or ambiguously associated, we use Gaussian-local evidence to supplement the instance prior rather than reduce it. %
The prior remains unchanged when $e_g(t)\le b_g(t)$. Otherwise, $s_g(t)$ moves from $b_g(t)$ toward $e_g(t)$ by the fraction $\rho_g$, so unreliable local evidence produces a smaller correction. %
Both terms aggregate the same query relevance function, evaluated on bank features and codebook prototypes, respectively (the two dashed arrows from the query in Fig.~\ref{fig:main_pipeline}). %

\paragraph{Multi-scale score and segmentation.}
We average the scores across semantic scales $\ell\in\mathcal S$ as $\bar s_g(t)=|\mathcal S|^{-1}\sum_{\ell\in\mathcal S}s_g^{(\ell)}(t)$, where $s_g^{(\ell)}(t)$ is Eq.~\ref{eq:fusion} evaluated at scale $\ell$.
In place of the relevance of aggregated features used by prior methods (\S\ref{sec:method_prelim}), we threshold $\bar s_g(t)$ and render the selected Gaussians to obtain the segmentation mask.
A new query changes the relevance scores $r(\mathbf x_o,t)$ and $r(\mathbf c_k,t)$; the representations, $\pi_g$, and $\rho_g$ are built once without text and shared by all queries.

\section{Experiments}
\label{sec:experiments}

\subsection{Experimental Setup}

\paragraph{Datasets and metrics.}
We evaluate open-vocabulary segmentation on LERF-OVS~\citep{kerr2023lerf,qin2024langsplat}, 3D-OVS~\citep{liu2023weakly}, and ScanNet~\citep{dai2017scannet}. We report mIoU on all three datasets, together with mAcc@$0.25$ on LERF-OVS and the class-averaged accuracy mAcc on ScanNet. %
LERF-OVS and 3D-OVS evaluate rendered masks; ScanNet evaluates 3D points in ten scenes following OpenGaussian~\citep{wu2024opengaussian}. Table~\ref{tab:lerf_main} separates methods that query rendered language
feature maps from those that score and select 3D Gaussians before rendering.
EviSplat and our LaGa baseline follow the latter.
For prior methods, we report mAcc@$0.25$ when available. %

\paragraph{Implementation.}
We use SAM~\citep{kirillov2023segment} for source-view mask generation and the OpenCLIP ViT-B/16 model of CLIP~\citep{radford2021learning} for image--text encoding across datasets.
On LERF-OVS we use three semantic scales, $D=32$, $K=256$ per scale, $10$k steps, and aggregation parameters $\beta=0.5$, $\gamma=0.25$, $T=0.07$. We use the same aggregation parameters across all four LERF-OVS scenes; the supplementary reports sensitivity analyses. %
Training and inference details are provided in the supplementary. %

\subsection{Main Results}

\begin{table}[t]
\centering
\footnotesize
\setlength{\tabcolsep}{2.6pt}
\caption{\textbf{LERF-OVS segmentation.}
$^{\dagger}$ denotes our reproduced baseline for LaGa~\citep{cen2025laga}; other prior results are cited as reported value. Metrics are four-scene averages, and our EviSplat reports three-run means $\pm$ population SD.
mAcc: mask IoU $>0.25$ success rate (\textbf{Best} / \underline{Second-Best}).}
\label{tab:lerf_main}
\resizebox{0.94\linewidth}{!}{%
\begin{tabular}{lccccrr}
\toprule
\textbf{Method} & \multicolumn{4}{c}{\textbf{Per-scene mIoU}} & \multicolumn{2}{c}{\textbf{Mean}} \\
\cmidrule(lr){2-5}\cmidrule(lr){6-7}
 & \textbf{Fig.} & \textbf{Tea.} & \textbf{Ram.} & \textbf{Wal.} & \textbf{mIoU} & \textbf{mAcc} \\
\midrule
\multicolumn{7}{l}{\textit{2D rendered-feature querying}} \\
LERF \citep{kerr2023lerf} & 38.6 & 45.0 & 28.2 & 37.9 & 37.4 & --  \\
LangSplat \citep{qin2024langsplat} & 44.7 & 65.1 & 51.2 & 44.5 & 51.4 & -- \\
Occam's LGS \citep{cheng2024occamlgs} & 58.6 & 70.2 & 51.0 & 65.3 & 61.3 & -- \\
OpenGaFF \citep{li2026opengaff} & \underline{64.3} & \textbf{76.1} & 53.8 & 65.8 & 65.0 & -- \\
CCL-LGS \citep{tian2025cclgs} & 61.2 & 71.8 & \underline{62.3} & \underline{67.1} & \underline{65.6} & -- \\
\addlinespace[1pt]\cdashline{1-7}[1pt/1.5pt]\addlinespace[2pt]
\multicolumn{7}{l}{\textit{3D Gaussian-level querying}} \\
OpenGaussian \citep{wu2024opengaussian} & 39.3 & 60.4 & 31.0 & 22.7 & 38.4 & 51.4 \\
LightSplat \citep{bang2026lightsplat} & 50.6 & 59.7 & 45.1 & 35.0 & 47.6 & 68.3 \\
OpenGaFF \citep{li2026opengaff} & 59.4 & 71.0 & 38.4 & 48.6 & 54.4 & 80.8 \\
VALA \citep{wang2025vala} & 60.4 & 70.6 & 45.4 & 55.7 & 58.0 & \underline{82.9} \\
LaGa \citep{cen2025laga} & 64.1 & 70.9 & 55.6 & 65.6 & 64.0 & -- \\
LaGa$^{\dagger}$ & 63.4 & 67.1 & 60.8 & 62.9 & 63.6 & 81.9 \\
\midrule
\textbf{EviSplat (Ours)} &
\textbf{74.8}\,{\scriptsize$\pm0.17$} &
\underline{73.5}\,{\scriptsize$\pm0.01$} &
\textbf{62.9}\,{\scriptsize$\pm0.27$} &
\textbf{71.5}\,{\scriptsize$\pm0.07$} &
\textbf{70.7}\,{\scriptsize$\pm0.10$} & \textbf{86.5}\,{\scriptsize$\pm0.00$} \\
\bottomrule
\end{tabular}
}
\end{table}

\paragraph{LERF-OVS.}
EviSplat achieves the highest mIoU on three of the four scenes and the highest mean, $70.7$, which is $5.1$ points above the best prior mean in either group (Tab.~\ref{tab:lerf_main}). It also reaches the highest mAcc@$0.25$, $86.5$ against $82.9$.
The largest margin over the strongest prior result is on \textsc{figurines}.

\begin{table}[t]
\centering
\caption{\textbf{Open-vocabulary segmentation on 3D-OVS and ScanNet.}
3D-OVS: five-scene mean mIoU. ScanNet: ten-scene point-level mIoU/mAcc under the OpenGaussian 19/15/10-class protocol.
Prior method results are cited from their original manuscripts (\textbf{Best} / \underline{Second-best}).}
\label{tab:other_datasets}
\vspace{2pt}
\begin{minipage}[t]{0.30\linewidth}
\centering
\textbf{(a) 3D-OVS}\par
\vspace{2pt}
\scriptsize
\setlength{\tabcolsep}{2.0pt}
\begin{tabularx}{\linewidth}{@{}Xr@{}}
\toprule
\textbf{Method} & \textbf{Mean} \\
\midrule
LangSplat \citep{qin2024langsplat} & 93.4 \\
ReasonGrounder \citep{liu2025reasongrounder} & 94.7 \\
Occam's LGS \citep{cheng2024occamlgs} & 95.0 \\
LaGa \citep{cen2025laga} & 95.3 \\
Hi-LSplat \citep{zhan2025hilsplat} & 96.1 \\
VLGS \citep{peng2025vlgs} & \underline{97.1} \\
\midrule
\textbf{EviSplat (Ours)} & \textbf{97.4} \\
\bottomrule
\end{tabularx}
\end{minipage}
\hfill
\begin{minipage}[t]{0.65\linewidth}
\centering
\textbf{(b) ScanNet}\par
\vspace{2pt}
\scriptsize
\setlength{\tabcolsep}{1.0pt}
\begin{tabularx}{\linewidth}{@{}X r@{\hspace{2pt}}r@{\hspace{7pt}}r@{\hspace{2pt}}r@{\hspace{7pt}}r@{\hspace{2pt}}r@{}}
\toprule
\textbf{Method}\,{\tiny(mIoU/mAcc)} & \multicolumn{2}{c}{\textbf{19 cls.}} & \multicolumn{2}{c}{\textbf{15 cls.}} & \multicolumn{2}{c}{\textbf{10 cls.}} \\
\midrule
OpenGaussian \citep{wu2024opengaussian} & 24.73 & 41.54 & 30.13 & 48.25 & 38.29 & 55.19 \\
ReLaGS \citep{xie2026relags} & 32.35 & 44.52 & \underline{40.04} & 60.59 & 47.17 & 66.08 \\
LaGa \citep{cen2025laga} & 32.50 & 49.10 & 35.50 & 53.50 & 42.60 & 63.20 \\
LUDVIG \citep{marrie2025ludvig} & 33.90 & 51.40 & 37.40 & 57.20 & 46.40 & 66.20 \\
LightSplat \citep{bang2026lightsplat} & 37.11 & \textbf{58.66} & 39.78 & \underline{60.91} & 47.78 & 68.21 \\
OpenInsGaussian \citep{huang2025openinsgaussian} & \underline{37.50} & 54.38 & 38.14 & 55.30 & \textbf{51.42} & 69.15 \\
\midrule
\textbf{EviSplat (Ours)} & \textbf{40.41} & \underline{57.88} & \textbf{44.85} & \textbf{62.89} & \underline{50.71} & \textbf{71.03} \\
\bottomrule
\end{tabularx}
\end{minipage}
\end{table}

\paragraph{3D-OVS and ScanNet.}
On 3D-OVS, EviSplat reaches $97.4$ mIoU, compared with $97.1$ for the best prior method.
On ScanNet, EviSplat achieves the highest mIoU for 19 and 15 classes and the second highest for 10 classes, together with the highest mAcc for 15 and 10 classes (Tab.~\ref{tab:other_datasets}).
These results extend the gains beyond the LERF-OVS scenes to both rendered-view and point-level segmentation.

\subsection{Component Analysis}

\begin{table}[t]
\centering
\small
\setlength{\tabcolsep}{2.5pt}
\caption{\textbf{Component and size analysis on LERF-OVS (mIoU).}
Baseline reproduces LaGa. $D$: latent dimension; $K$: prototypes per semantic scale.
Local storage totals four scenes and excludes shared components.}
\label{tab:lerf_component}
\begin{tabular}{lcccccc}
\toprule
 & \textbf{Fig.} & \textbf{Tea.} & \textbf{Ram.} & \textbf{Wal.} & \textbf{Mean} & \shortstack{\textbf{Local storage}\textbf{(MiB)} $\boldsymbol{\downarrow}$} \\
\midrule
\multicolumn{7}{l}{\textit{Instance observations and the evidence field}} \\
Baseline~\citep{cen2025laga} & 63.4 & 67.1 & 60.8 & 62.9 & 63.6  & -- \\
Query-guided evidence aggregation (instance only) & 65.2 & 69.8 & 65.1 & 64.8 & 66.2  & -- \\
\midrule
\multicolumn{7}{l}{\textit{Evidence representation and aggregation}} \\
Mean observation feature & 67.3 & 72.8 & \textbf{65.6} & 67.3 & 68.2  & -- \\
Mean prototype relevance (probability-weighted) & 66.6 & 72.6 & 65.4 & 65.5 & 67.5  & -- \\
Maximum prototype relevance & 31.8 & 49.4 & 37.1 & 58.5 & 44.2  & -- \\
\midrule
\multicolumn{7}{l}{\textit{Representation size and storage}} \\
Sparse empirical distribution & 73.5 & 72.7 & 63.0 & 71.4 & 70.1 & 307.7 \\
EviSplat ($D=8,K=128$) & 70.9 & 70.0 & 63.3 & 67.1 & 67.8  & \textbf{86.6} \\
EviSplat ($D=8,K=256$) & 71.5 & 73.2 & 63.8 & 68.6 & 69.3  & 86.7 \\
EviSplat ($D=16,K=256$) & 72.5 & \textbf{73.7} & 63.5 & 70.9 & 70.2  & 173.3 \\
EviSplat ($D=32,K=256$) & \textbf{74.8} & 73.5 & 62.9 & \textbf{71.5} & \textbf{70.7}  & 346.7 \\
\bottomrule
\end{tabular}
\end{table}

\paragraph{Instance observations and the evidence field.}
Across variants, we keep the reconstructed geometry, LaGa affinity field, renderer, and post-processing fixed.
The instance-only variant scores each Gaussian by the instance prior $b_g(t)$ alone (Eq.~\ref{eq:instance_prior}), without the Gaussian-local correction; it raises the baseline from $63.6$ to $66.2$ mIoU without additional representation learning (Tab.~\ref{tab:lerf_component}), already above the best prior mean of $65.6$ (Tab.~\ref{tab:lerf_main}).
Averaging the query relevance scores of all retained observations yields $62.8$ mIoU and taking the maximum score yields $60.0$, compared with $66.2$ for the top-$\beta$ mean, supporting query-dependent selection of a subset of observations. %
Adding Gaussian-local evidence with reliability-weighted fusion yields a further $4.5$-point gain, reaching $70.7$ mIoU. It improves three scenes over the instance-only variant but lowers \textsc{ramen} from $65.1$ to $62.9$.%

\paragraph{Evidence representation and aggregation.}
We also vary how Gaussian-local evidence is represented and aggregated, using $\beta=0.5$ for the instance prior and keeping the fusion rule fixed (Tab.~\ref{tab:lerf_component}).
The \emph{mean observation feature} variant averages each Gaussian's original CLIP observation features with their observation weights, normalizes the result, and scores this single feature against the query, yielding $68.2$ mIoU.
The other two alternatives use the same learned evidence distributions as EviSplat: averaging all prototype relevances with their probabilities as weights yields $67.5$, while taking the maximum prototype relevance regardless of its probability yields $44.2$. This maximum assigns the same local score to every Gaussian at a given scale and query.
Our aggregation over a fixed probability mass achieves $70.7$ mIoU averaged over three runs.
These results support retaining multiple appearances and considering their probability support during aggregation, although the mean observation feature performs better on \textsc{ramen}.

\paragraph{Representation size and storage.}
We use $D=32$ as the default for its higher mean accuracy.
At $K=256$, mean mIoU decreases as the latent dimension is reduced; at $D=8$, reducing $K$ from $256$ to $128$ further lowers it from $69.3$ to $67.8$ (Tab.~\ref{tab:lerf_component}).
At $\beta=0.5$, the $D=16$ field provides a lower-storage operating point: it achieves comparable mean mIoU to directly stored sparse empirical distributions ($70.2$ vs.\ $70.1$), while reducing local representation storage by $43.7\%$ ($173.3$ vs.\ $307.7$ MiB).
With the full-view observation bank and codebooks fixed, increasing local evidence coverage from $10\%$ to $100\%$ raises sparse storage from $121.6$ to $307.7$ MiB, while the $D=16$ field remains at $173.3$ MiB.
These storage measurements exclude the observation bank and other shared components; the supplementary provides the view-coverage analysis and additional training and query costs.

\begin{figure}[t]
\centering
\includegraphics[width=\linewidth]{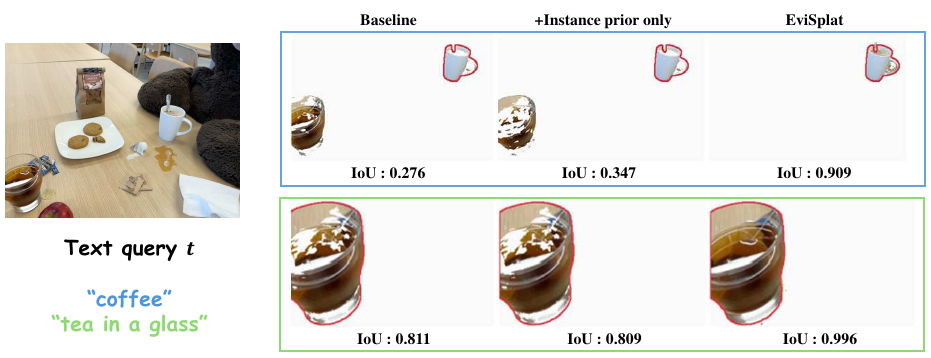}
\caption{\textbf{Qualitative comparison on LERF-OVS.}
Left: the input view and color-coded queries. Right: the reproduced baseline, the instance prior alone, and EviSplat.}
\label{fig:qualitative}
\end{figure}

\paragraph{Qualitative comparison.}
Fig.~\ref{fig:qualitative} compares the baseline, the instance-only variant, and EviSplat for three queries in the same view of \textsc{teatime}.
For ``coffee,'' the instance-only variant raises IoU from $0.276$ to $0.347$ but still includes part of the glass; EviSplat more closely matches the annotated mug, reaching $0.909$.
For ``tea in a glass,'' the instance-only masks remain similar to the baseline, whereas EviSplat provides more complete coverage of the annotated targets.
Local evidence improves segmentation even when instance-only gains are small.

\section{Conclusion}
\label{sec:conclusion}

We presented EviSplat, an approach to open-vocabulary 3D segmentation that preserves individual observation features for query-dependent selection.
Instance observation banks retain multi-view evidence, while a Gaussian evidence field learns a compact distribution over the visual appearances supported at each Gaussian.
Query-guided evidence aggregation forms an instance prior from the most relevant observations and refines each Gaussian's score with reliable Gaussian-local evidence.
This allows different queries to draw on different cues from the same preserved evidence, without retraining the representation.
Across three datasets, EviSplat improves over the corresponding baselines under both rendered-view and point-level evaluation protocols, with component analyses showing complementary gains from instance observation selection and Gaussian-local evidence.
\paragraph{Limitations.}
EviSplat remains dependent on source-mask quality and frozen CLIP features; preserving observations alone may not overcome CLIP's limitations in understanding attribute combinations and spatial relations~\citep{yuksekgonul2023bags}.
Refining the fixed Gaussian-to-instance assignments and improving compositional query understanding remain directions for future work.

\subsection*{AI use statement}
We used generative AI tools to assist with text editing and proofreading, implementing and debugging experimental and analysis code, and formatting schematic figure drafts.
Additional support was limited to structural layout exploration and verifying equation formatting for clarity.
All numerical values, experimental designs, mathematical derivations, and literature citations were independently verified and validated using original sources and code execution outputs.
The authors retain full responsibility for all content, scientific claims, experimental results, and final text presented in this manuscript.
The authors take responsibility for the final manuscript, experimental results, code, and figures, including all content developed with AI assistance.

\subsection*{Ethics statement}
This work studies open-vocabulary 3D segmentation using existing research benchmarks.
The method may inherit biases and errors from pretrained vision--language and segmentation models.
In particular, segmentation quality can vary with the query, object appearance, and visibility, and benchmark performance does not establish reliability in safety-critical applications.
Language-based retrieval in reconstructed indoor scenes can also expose sensitive information about personal spaces and belongings.
Applications involving such scenes should account for consent, access control, and privacy when collecting, querying, or sharing the data.
Use and redistribution of benchmark data and pretrained models should follow their respective licenses and access conditions.

\subsection*{Reproducibility statement}
Section~\ref{sec:method} describes the evidence representation, training objective, and query-guided evidence aggregation.
Section~\ref{sec:experiments} describes the datasets and evaluation settings.
Appendix~\ref{app:implementation} specifies observation preparation, weight normalization and empty-target handling, and training and inference conventions.
Appendix~\ref{app:protocol} explains metric averaging and the repeated-run protocol, including which components are held fixed when varying the evidence-field training seed.
Appendix~\ref{app:additional} provides controlled comparisons of evidence representations and aggregation rules, together with sensitivity and observation-quality experiments.
These descriptions identify the configurations and controls needed to interpret the comparisons and distinguish single-run ablations from repeated-run results.

\bibliography{iclr2026_conference}
\bibliographystyle{iclr2027_conference}

\clearpage
\appendix
\counterwithin{table}{section}
\counterwithin{figure}{section}
\counterwithin{equation}{section}
\section{Implementation and Complexity Details}
\label{app:implementation}

This appendix provides implementation and evaluation details, followed by analyses of observation selection, Gaussian-local evidence, and representation size.

\subsection{Observation Preparation and Training}

\begin{table}[t]
\centering
\small
\setlength{\tabcolsep}{4pt}
\caption{\textbf{Implementation and evaluation settings.}}
\label{tab:app_settings}
\begin{tabular}{llll}
\toprule
\textbf{Setting} & \textbf{LERF-OVS} & \textbf{3D-OVS} & \textbf{ScanNet} \\
\midrule
\multicolumn{4}{l}{\emph{Observations}} \\
Mask generator & \multicolumn{3}{c}{SAM ViT-H} \\
Image encoder & \multicolumn{3}{c}{OpenCLIP ViT-B/16; 512-D unit-norm features} \\
Minimum mask size & $>400$ px & $>400$ px & $>400$ px \\
\midrule
\multicolumn{4}{l}{\emph{Evidence field}} \\
Semantic scales & $3$ & $1$ & $1$ \\
Latent dimension $D$ & $32$ & $32$ & $32$ \\
Codebook size $K$ (per scale) & $256$ & $128$ & $256$ \\
Optimization steps & $10{,}000$ & $10{,}000$ & $10{,}000$ \\
Latent optimizer & \multicolumn{3}{c}{SparseAdam, learning rate $3\times10^{-3}$, constant} \\
Decoder optimizer & \multicolumn{3}{c}{AdamW, learning rate $10^{-3}$, weight decay $10^{-4}$, cosine decay} \\
Trainable parameters & \multicolumn{3}{c}{Gaussian latents and scale-specific linear decoders only} \\
\midrule
\multicolumn{4}{l}{\emph{Query-guided aggregation}} \\
Observation fraction $\beta$ & $0.5$ & $0.5$ & $0.5$ \\
Probability mass $\gamma$ & $0.25$ & $0.25$ & $0.25$ \\
Instance assignment & Soft, $T=0.07$ & Soft, $T=0.07$ & Soft, $T=0.07$ \\
CLIP relevance scale & \multicolumn{3}{c}{$\lambda=10$ against the fixed canonical negatives} \\
\midrule
\multicolumn{4}{l}{\emph{Evaluation}} \\
Evaluation unit & \multicolumn{2}{c}{Rendered views} & Points \\
Score normalization & \multicolumn{2}{c}{Quantile clip $0.25$, then min--max} & Raw fused scores \\
Cosine gate & $0.23$ & $0$ (sign of cosine) & Off \\
Selection threshold & $0.7$ (Gaussian) & $0.7$ (Gaussian) & Zero-score rejection \\
Mask threshold & $0.3$ & $0.8$ (rendered) & n/a \\
Score smoothing & None & None & None \\
Class subsets & n/a & n/a & $19/15/10$, scored separately \\
\midrule
\multicolumn{4}{l}{\emph{Runs}} \\
Shared components & \multicolumn{3}{c}{Fixed 3DGS and affinity-field checkpoints for within-method ablations} \\
Hardware & \multicolumn{3}{c}{NVIDIA B200, CUDA 12.8} \\
Software & \multicolumn{3}{c}{PyTorch 2.8.0, gsplat 1.3.0, open\_clip 3.3.0} \\
\bottomrule
\end{tabular}
\end{table}

\paragraph{Observation filtering.}
For LERF-OVS, both the instance observation bank and the evidence-set association cache retain masks with more than $400$ pixels.
The bank reuses the baseline's cached mask filtering and instance memberships; observations labeled as noise by HDBSCAN belong to no bank.
For the association cache, masks at each scale are compared with masks at earlier, finer scales in the same view. A mask is removed if its IoU with any such mask is at least $0.8$; a filtering step is skipped if it would remove every remaining mask at that scale.
The bank and association cache therefore use different filtered subsets of the observation set $\mathcal O$ defined in the main text. Codebook construction uses the subset retained in the association cache.

\paragraph{Observation weights and empty targets.}
We normalize assignment confidence within each source view so that views with many associated Gaussians do not automatically receive more total weight.
At a fixed semantic scale, let $\mathcal G_v$ contain the Gaussians with a selected observation from view $v$.
The implementation uses
\begin{equation}
S_v=\sum_{h\in\mathcal G_v}\omega_h(o_h^{(v)}),
\qquad
w_g(o_g^{(v)})=\frac{\omega_g(o_g^{(v)})}{\max(S_v,\varepsilon)},
\qquad \varepsilon=10^{-12}.
\label{eq:app_observation_weights}
\end{equation}
Thus, the weights sum to one when $S_v\geq\varepsilon$. If all assignment margins in a view are zero, that view contributes zero weight; no uniform fallback is added.
For each Gaussian, define $Z_g=\sum_{o\in\mathcal E_g}w_g(o)$.
Equation~\ref{eq:exact_target} defines a normalized target only when $Z_g>0$.
When $Z_g=0$, either because the evidence set is empty or because all its weights are zero, the implementation stores an all-zero target row as a placeholder, not as a probability distribution.
At each training step, a semantic scale is sampled uniformly, followed by Gaussians sampled with probability proportional to $Z_g$ at that scale. Zero-weight rows are therefore never sampled and contribute no loss; training stops with an error if a scale has no positive-weight rows.
This exclusion applies to a Gaussian at that scale: its shared latent may still be trained through other scales with positive weight.
For empty evidence or all-zero assignment margins, observation reliability is zero, so fusion retains the instance prior regardless of the decoded local distribution.

\paragraph{Optimization.}
The evidence field is trained to reproduce the empirical evidence distributions before any text query is given.
Gaussian latents are implemented as a sparse lookup table. We use SparseAdam, a sparse-gradient variant of Adam~\citep{kingma2015adam},
for the latents and AdamW~\citep{loshchilov2019decoupled}
for the scale-specific decoders, with settings listed in
Tab.~\ref{tab:app_settings}. %
Only the decoder learning rate follows cosine decay; the latent learning rate remains constant.
At query time, the stored observation features and prototypes are reused without another image-encoder pass.

\subsection{Scoring Implementation and Post-processing}

This section specifies numerical conventions and computation shared across queries for the scoring terms defined in \S\ref{sec:method_query_scoring}, together with the post-processing settings.

\paragraph{Numerical conventions.}
The relative assignment margin is computed as the difference between the largest and second-largest association values divided by the largest value clamped below at $\varepsilon=10^{-12}$, then clipped to $[0,1]$. A view with only one candidate mask uses margin one for its associated Gaussians. Per-view weight normalization follows Eq.~\ref{eq:app_observation_weights}.
For the relevance function in Eq.~\plaineqref{eq:relevance}, the implementation multiplies both the query and canonical-negative similarity logits by $\lambda=10$ before exponentiation.

\paragraph{Shared scoring computation.}
Soft assignments and observation reliability are computed once and reused across queries.
For Gaussian-local evidence, prototype relevances are sorted once per query and scale; this order is shared by all Gaussians when accumulating the probability mass specified in Eq.~\plaineqref{eq:gaussian_local_evidence}.

\paragraph{Post-processing.}
For LERF-OVS, we retain LaGa's post-processing and apply it to the fused Gaussian scores. After averaging across semantic scales, scores below the $25$th percentile are clipped to that percentile, followed by min--max normalization. For the inherited cosine gate, each Gaussian uses its highest-affinity instance at each scale. Within that instance, we take the query cosine similarity of the LaGa descriptor selected by its weighted relevance, then take the maximum across scales. Fused scores are set to zero when this value is below $0.23$. We then select Gaussians with scores above $0.7$, render the selection, and threshold the rendered mask at $0.3$. No additional 3D smoothing is applied.

\section{Evaluation Protocol}
\label{app:protocol}

\paragraph{Averaging.}
On LERF-OVS and 3D-OVS, the mIoU of a scene is the mean IoU over all annotated (frame, class) pairs rather than the mean of per-class averages.
Unequal numbers of annotated frames per class can make these averages differ; we use the same convention for all rows within each comparison.
The reported mean over scenes is unweighted.

\paragraph{Single operating point.}
Every reported configuration uses one set of aggregation hyperparameters ($\beta$, $\gamma$, $T$) and one set of post-processing thresholds across all four LERF-OVS scenes.
Table~\ref{tab:app_settings} summarizes the training and inference settings used for the reported results.

\paragraph{Repeated runs.}
The three-run statistics in Tab.~\ref{tab:lerf_main} and the $D=32,K=256$ row of Tab.~\ref{tab:lerf_component} vary the seed of evidence-field training (codebook initialization, Gaussian sampling, and latent initialization) while keeping the photometric scene and the affinity field fixed.
Within-method ablations reuse the same affinity-field checkpoints to isolate changes to evidence representation and aggregation.

\paragraph{LERF-OVS accuracy.}
For each scene, mAcc@$0.25$ is the fraction of annotated (frame, query) pairs whose mask IoU exceeds $0.25$; the reported mean averages the four scenes.
It is distinct from per-class point recall on ScanNet and from localization accuracy based on the highest-scoring image pixel.
For the main setting, all three evidence-field seeds give $86.54\%$: their masks differ, but the numbers of pairs crossing the IoU threshold are unchanged.

\section{Additional Experiments on LERF-OVS}
\label{app:additional}

The following experiments examine the roles of observation selection, Gaussian-local evidence, and learned representation size in the accuracy gains reported in the main paper.
We also test sensitivity to observation quality and examine how cross-view appearance differences relate to the gains.
Unless varied explicitly, we use the main LERF-OVS setting $(\beta,\gamma,T)=(0.5,0.25,0.07)$.
The aggregation and sensitivity ablations use one of the three evidence-field seeds reported in Tab.~\ref{tab:lerf_main}, with its checkpoint fixed across inference variants.

\paragraph{Full comparison on LERF-OVS.}
Tab.~\ref{tab:app_lerf_full} extends the main comparison with additional methods and per-scene accuracies. The upper group queries rendered feature maps; the lower group selects Gaussians and renders the selected geometry. We report mask success rates only for the latter protocol.

\begin{table}[t]
\centering
\footnotesize
\setlength{\tabcolsep}{2.4pt}
\caption{\textbf{Extended LERF-OVS comparison.} $\dagger$: reproduced by us and used as our baseline; $\ddagger$: re-evaluated by LightSplat under Gaussian-level querying. Other prior results are as reported. EviSplat means and standard deviations follow Tab.~\ref{tab:lerf_main}. Bold/underline: best/second-best displayed mean per column; ties share emphasis.}
\label{tab:app_lerf_full}
\resizebox{\linewidth}{!}{%
\begin{tabular}{lcccc|cccc|cc}
\toprule
 & \multicolumn{4}{c|}{\textbf{mIoU}} & \multicolumn{4}{c|}{\textbf{mAcc@$0.25$}} & \multicolumn{2}{c}{\textbf{Mean}} \\
\cmidrule(lr){2-5}\cmidrule(lr){6-9}\cmidrule(lr){10-11}
\textbf{Method} & Fig. & Tea. & Ram. & Wal. & Fig. & Tea. & Ram. & Wal. & mIoU & mAcc \\
\midrule
\multicolumn{11}{l}{\textit{2D rendered-feature querying}} \\
LERF \citep{kerr2023lerf}  & 38.6  & 45.0  & 28.2  & 37.9  & \textcolor{gray}{--}  & \textcolor{gray}{--}  & \textcolor{gray}{--}  & \textcolor{gray}{--}  & 37.4  & \textcolor{gray}{--} \\
LEGaussians \citep{shi2024legaussians}  & 60.3  & 44.5  & 52.6  & 41.4  & \textcolor{gray}{--}  & \textcolor{gray}{--}  & \textcolor{gray}{--}  & \textcolor{gray}{--}  & 46.9  & \textcolor{gray}{--} \\
LangSplat \citep{qin2024langsplat}  & 44.7  & 65.1  & 51.2  & 44.5  & \textcolor{gray}{--}  & \textcolor{gray}{--}  & \textcolor{gray}{--}  & \textcolor{gray}{--}  & 51.4  & \textcolor{gray}{--} \\
Occam's LGS \citep{cheng2024occamlgs}  & 58.6  & 70.2  & 51.0  & 65.3  & \textcolor{gray}{--}  & \textcolor{gray}{--}  & \textcolor{gray}{--}  & \textcolor{gray}{--}  & 61.3  & \textcolor{gray}{--} \\
VLGS \citep{peng2025vlgs}  & 58.1  & 73.5  & 61.4  & 54.8  & \textcolor{gray}{--}  & \textcolor{gray}{--}  & \textcolor{gray}{--}  & \textcolor{gray}{--}  & 62.0  & \textcolor{gray}{--} \\
ReLaGS \citep{xie2026relags}  & \underline{64.7}  & \textbf{81.0}  & 51.2  & 60.6  & \textcolor{gray}{--}  & \textcolor{gray}{--}  & \textcolor{gray}{--}  & \textcolor{gray}{--}  & 64.4  & \textcolor{gray}{--} \\
OpenGaFF \citep{li2026opengaff}  & 64.3  & \underline{76.1}  & 53.8  & 65.8  & \textcolor{gray}{--}  & \textcolor{gray}{--}  & \textcolor{gray}{--}  & \textcolor{gray}{--}  & 65.0  & \textcolor{gray}{--} \\
CCL-LGS \citep{tian2025cclgs}  & 61.2  & 71.8  & \underline{62.3}  & \underline{67.1}  & \textcolor{gray}{--}  & \textcolor{gray}{--}  & \textcolor{gray}{--}  & \textcolor{gray}{--}  & \underline{65.6}  & \textcolor{gray}{--} \\
\midrule
\multicolumn{11}{l}{\textit{3D Gaussian-level querying}} \\
LangSplat$^{\ddagger}$ \citep{qin2024langsplat}  & 9.6  & 7.9  & 3.5  & 9.6  & 14.3  & 8.5  & 5.6  & 9.1  & 7.7  & 9.4 \\
LEGaussians$^{\ddagger}$ \citep{shi2024legaussians}  & 16.4  & 21.9  & 14.2  & 17.2  & 25.0  & 33.9  & 28.2  & 27.3  & 17.4  & 28.6 \\
SuperGSeg \citep{liang2024supergseg}  & 43.7  & 55.3  & 18.1  & 26.7  & 60.7  & 78.0  & 23.9  & 45.5  & 35.9  & 52.0 \\
OpenGaussian \citep{wu2024opengaussian}  & 39.3  & 60.4  & 31.0  & 22.7  & 55.4  & 76.3  & 42.3  & 31.8  & 38.4  & 51.4 \\
OpenInsGaussian \citep{huang2025openinsgaussian}  & 53.8  & 58.6  & 26.4  & 31.8  & 76.8  & 78.0  & 43.7  & 50.0  & 42.6  & 62.1 \\
Dr.~Splat \citep{kim2025drsplat}  & 54.4  & 57.4  & 24.3  & 37.1  & 80.4  & 78.0  & 35.2  & 63.6  & 43.3  & 64.3 \\
Occam's LGS \citep{cheng2024occamlgs}  & 52.9  & 61.0  & 32.0  & 43.0  & 78.6  & \textbf{93.2}  & 54.9  & 72.7  & 47.2  & 74.8 \\
LightSplat \citep{bang2026lightsplat}  & 50.6  & 59.7  & 45.1  & 35.0  & 76.8  & 79.7  & 57.8  & 59.1  & 47.6  & 68.3 \\
OpenGaFF \citep{li2026opengaff}  & 59.4  & 71.0  & 38.4  & 48.6  & \textbf{92.9}  & \underline{89.8}  & 63.4  & 77.3  & 54.4  & 80.8 \\
VALA \citep{wang2025vala}  & 60.4  & 70.6  & 45.4  & 55.7  & 89.3  & 88.1  & \underline{67.6}  & \underline{86.4}  & 58.0  & \underline{82.9} \\
LaGa \citep{cen2025laga} & 64.1 & 70.9 & 55.6 & 65.6 & \textcolor{gray}{--} & \textcolor{gray}{--} & \textcolor{gray}{--} & \textcolor{gray}{--} & 64.0 & \textcolor{gray}{--} \\
LaGa$^{\dagger}$  & 63.4  & 67.1  & 60.8  & 62.9  & 78.6  & 86.4  & \textbf{76.1}  & \underline{86.4}  & 63.6  & 81.9 \\
\midrule
\textbf{EviSplat (Ours)}  & \textbf{74.8}\,{\scriptsize$\pm0.17$}  & 73.5\,{\scriptsize$\pm0.01$}  & \textbf{62.9}\,{\scriptsize$\pm0.27$}  & \textbf{71.5}\,{\scriptsize$\pm0.07$}  & \underline{91.1}  & 88.1  & \textbf{76.1}  & \textbf{90.9}  & \textbf{70.7}\,{\scriptsize$\pm0.10$}  & \textbf{86.5}\,{\scriptsize$\pm0.00$} \\
\bottomrule
\end{tabular}
}
\end{table}

\subsection{Observation Quality}

\begin{figure}[t]
\centering
\begin{minipage}[t]{0.48\linewidth}
\centering
\includegraphics[width=\linewidth]{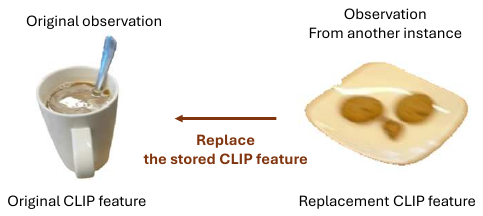}
\par\smallskip\small (a) Feature replacement
\end{minipage}\hfill
\begin{minipage}[t]{0.48\linewidth}
\centering
\includegraphics[width=\linewidth]{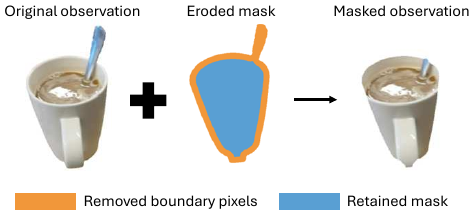}
\par\smallskip\small (b) Mask erosion
\end{minipage}
\caption{\textbf{Controlled changes to source observations.} (a) Another instance supplies the replacement CLIP feature; the source RGB and mask stay fixed. (b) Boundary pixels are removed before CLIP re-encoding. The illustrated erosion is exaggerated for visibility; orange and blue indicate removed and retained mask pixels, respectively.}
\label{fig:app_observation_quality}
\end{figure}

We test sensitivity to misleading features and incomplete masks using feature replacement and mask erosion (Fig.~\ref{fig:app_observation_quality}).
For both experiments, we perturb nested subsets of $10\%$ and $30\%$ of source views, keeping geometry, grouping, and codebooks fixed. The affected representations are rebuilt and the evidence field is retrained; all three variants receive the same modified observations.
The rates refer to selected source views, not the removed mask area or fraction of all feature vectors.
Table~\ref{tab:app_observation_quality} reports mean mIoU and changes from each experiment's control. The matched local runs are distinct from the baseline run in the main table.
These tests concern source-observation perturbations on a fixed scene, rather than physical occlusion.

\paragraph{Feature replacement.}
To test sensitivity to mismatched visual features, we replace a stored CLIP feature with one from another instance, leaving the source RGB and mask unchanged (Fig.~\ref{fig:app_observation_quality}a).
Replacement features come from the same scene and semantic scale, with non-overlapping fixed-instance memberships; they are not guaranteed to belong to a different semantic class.
EviSplat is more robust than the baseline to the tested replacements: at $30\%$ replacement, its mIoU drops by $1.91$ points, compared with $4.78$ points for the baseline and $2.07$ points for instance-only.

\paragraph{Mask erosion.}
To test sensitivity to incomplete masks, we shrink the source mask and re-encode the remaining crop (Fig.~\ref{fig:app_observation_quality}b).
The erosion radius is $\max(1,\operatorname{round}(0.03\sqrt{A/\pi}))$ pixels for mask area $A$; masks that would become empty retain their original shape.
Existing observation slots are kept, and renderer associations are recomputed for affected views.
Both erosion conditions and their control re-encode the same $30\%$ view subset using the same encoder and crop procedure. The $10\%$ condition erodes only a nested subset, while the control performs no erosion.
This separates mask shrinkage from changes caused by re-encoding.
EviSplat maintains nearly unchanged accuracy, with changes within $0.13$ points of this control; all three variants change by at most $0.60$ points.

\begin{table}[t]
\centering\small
\setlength{\tabcolsep}{5pt}
\caption{\textbf{Sensitivity to source-observation perturbations.} Four-scene mean mIoU (\%). Parentheses give changes in percentage points from the control within each block. Erosion uses a matched re-encoding control. Light-blue shading highlights changes from the corresponding control.}
\label{tab:app_observation_quality}
\begin{tabular}{lrrr}
\toprule
Perturbed source views & Baseline & Instance-only & EviSplat \\
\midrule
\multicolumn{4}{l}{\emph{Feature replacement}} \\
Clean & $64.15$ & $66.22$ & $70.54$ \\
10\% & $62.48$ \colorbox{cyan!10}{$(-1.67)$} & $66.30$ \colorbox{cyan!10}{$(+0.08)$} & $70.22$ \colorbox{cyan!10}{$(-0.32)$} \\
30\% & $59.38$ \colorbox{cyan!10}{$(-4.78)$} & $64.15$ \colorbox{cyan!10}{$(-2.07)$} & $68.63$ \colorbox{cyan!10}{$(-1.91)$} \\
\midrule
\multicolumn{4}{l}{\emph{Mask erosion}} \\
Re-encoding control & $62.91$ & $66.42$ & $70.64$ \\
10\% & $63.50$ \colorbox{cyan!10}{$(+0.60)$} & $66.90$ \colorbox{cyan!10}{$(+0.48)$} & $70.77$ \colorbox{cyan!10}{$(+0.13)$} \\
30\% & $63.13$ \colorbox{cyan!10}{$(+0.22)$} & $66.35$ \colorbox{cyan!10}{$(-0.07)$} & $70.61$ \colorbox{cyan!10}{$(-0.03)$} \\
\bottomrule
\end{tabular}
\end{table}

\paragraph{Test-view occlusion.}
We test whether EviSplat also improves segmentation when other objects hide part of the target, as with ``green apple'' in the \textsc{figurines} example of Fig.~\ref{fig:app_qualitative}.
We inspect all $208$ frame--query pairs from the four LERF-OVS scenes and estimate how much of each target is hidden, grouping cases into $\leq30\%$ and $>30\%$ occlusion.
The labels combine author annotations and visual estimates. Cases with unclear occlusion or targets cut off by the image boundary are excluded. The final analysis includes $125$ pairs with $\leq30\%$ occlusion and $19$ pairs with $>30\%$ occlusion.
We compare the existing baseline, instance-only, and EviSplat predictions on the same pairs, without changing the source observations or retraining.
Table~\ref{tab:app_occlusion} shows that EviSplat improves mean IoU over the baseline in both groups, with a larger gain for targets with more than $30\%$ occlusion ($10.42$ vs.\ $5.52$ points).
The gains vary across scenes, and the higher-occlusion group in \textsc{teatime} contains only one pair. Each frame--query pair has equal weight in these averages, unlike the scene-level averaging in the main table.

\begin{table}[t]
\centering\small
\setlength{\tabcolsep}{5pt}
\caption{\textbf{Segmentation under estimated test-view occlusion.} Mean IoU (\%) over frame--query pairs pooled across four LERF-OVS scenes. Groups use provisional visual estimates of external occlusion. $n$ is the number of pairs; shaded values give EviSplat's gain over the baseline in percentage points.}
\label{tab:app_occlusion}
\begin{tabular}{lrrrrr}
\toprule
Estimated occlusion & $n$ & Baseline & Instance-only & EviSplat & Gain \\
\midrule
$\leq30\%$ & 125 & 66.04 & 69.00 & \textbf{71.56} & \cellcolor{cyan!10}$+5.52$ \\
$>30\%$ & 19 & 55.98 & 58.11 & \textbf{66.40} & \cellcolor{cyan!10}$+10.42$ \\
\bottomrule
\end{tabular}
\end{table}

\subsection{Query-Guided Evidence Aggregation}

We first compare averaging observation features with query-guided observation selection, then examine Gaussian-local evidence, fusion, parameter sensitivity, and variation across evidence-field training seeds.

\paragraph{Averaging features versus selecting observations.}
We test whether keeping observations available for query-guided selection helps distinguish a target from competing instances.
Using the same observation bank and CLIP relevance function, we compare averaging features before scoring, averaging scores from individual observations, and averaging the top $50\%$ of observation scores.
The averaged feature is L2-normalized before scoring; all three methods use the same text features and canonical negatives.
We measure the \emph{score gap}: the target's relevance minus the highest relevance among competing instances. A positive gap places the target above all competitors.
Table~\ref{tab:app_mean_feature} shows that selection increases the mean gap from $0.065$ to $0.078$, with a positive mean improvement in each of the four scenes.
This comparison supports retaining observations for query-guided selection rather than summarizing them by a single mean feature.

\begin{table}[t]
\centering\small
\caption{\textbf{Target separation using the same observation bank.} Mean target--competitor score gap ($\uparrow$), averaged within each scene--query pair and then across $66$ pairs. Ground-truth-matched instances and scales with mask IoU $\geq0.5$ retain $202$ of $208$ frame--query pairs. Competitors have zero target-mask overlap and at least $\max(16,10^{-4}HW)$ rendered pixels for image size $H\times W$. All methods use the same competitor set. Values are relevance gaps, not segmentation IoU.}
\label{tab:app_mean_feature}
\begin{tabular}{lr}
\toprule
Observation aggregation & Mean score gap $\uparrow$ \\
\midrule
Average features, then score & $0.065$ \\
Score observations, then average & $0.054$ \\
Score observations, then select top $50\%$ & $\mathbf{0.078}$ \\
\bottomrule
\end{tabular}
\end{table}

\begin{figure}[t]
\centering
\begin{minipage}[c]{0.20\linewidth}
\centering
\includegraphics[width=\linewidth]{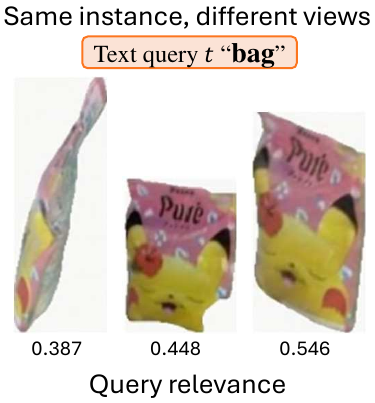}
\par\smallskip\small (a) Query: ``bag''
\end{minipage}\hfill
\begin{minipage}[c]{0.78\linewidth}
\centering
\includegraphics[width=\linewidth]{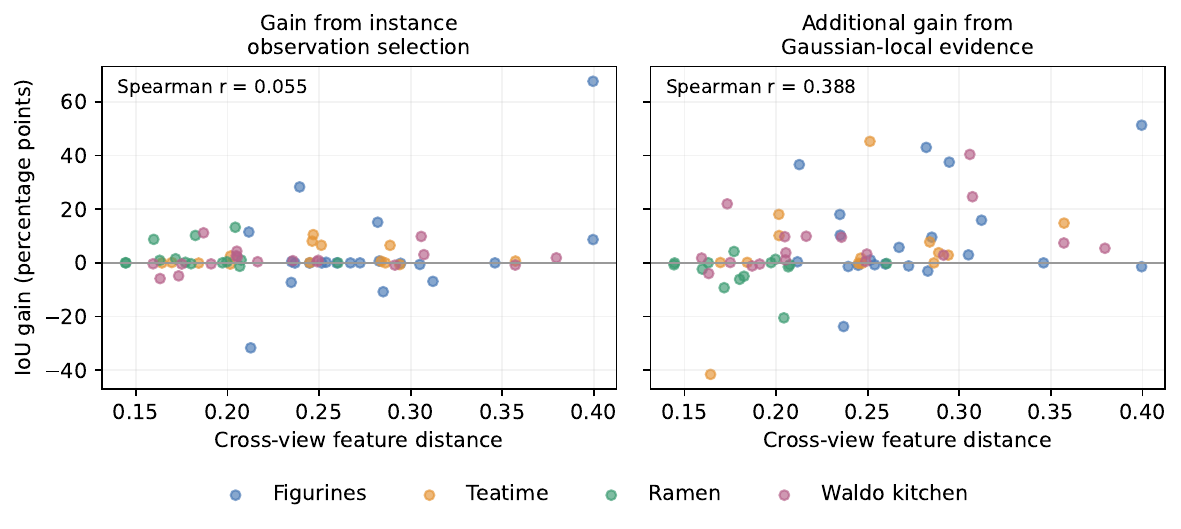}
\par\smallskip\small (b) Appearance variation and IoU gains
\end{minipage}
\caption{\textbf{Appearance variation and Gaussian-local evidence.} (a) Views of the same bag with query relevance scores. (b) Cross-view feature distance versus IoU gains from the instance prior (left plot) and additional Gaussian-local evidence (right plot), for all $67$ scene--query pairs. Feature distance is distinct from query relevance; positive gains indicate improved IoU.}
\label{fig:app_appearance}
\end{figure}

\paragraph{Appearance variation and Gaussian-local evidence.}
Different views of the same instance can reveal different visual cues, as illustrated by the bag observations in Fig.~\ref{fig:app_appearance}(a). The instance prior aggregates observations at the instance level, while Gaussian-local evidence additionally accounts for the appearance support associated with each Gaussian.
We examine whether greater variation across observed appearances is associated with larger gains from this local refinement.
We measure appearance differences using cosine distances between unit CLIP observation features from different views, averaging observation pairs within each view pair and then weighting view pairs equally.
Larger distances indicate greater differences between observed appearances. This measure does not use text relevance.
We select the instance and scale by ground-truth correspondence and average measurements and IoU gains across frames within each of the $67$ scene--query pairs.

The two plots in Fig.~\ref{fig:app_appearance}(b) separate the gain from the instance prior (left plot) from the additional gain of the full model (right plot).
The horizontal axis measures cross-view feature differences. The vertical axis shows the IoU gain of the instance prior over the baseline in the left plot, and of the full model over the instance prior in the right plot. Points above zero indicate an improvement.
Cross-view feature distance is more strongly associated with the additional gain from Gaussian-local evidence (Spearman $r=0.388$) than with the gain from instance observation selection ($r=0.055$).
Thus, larger observed feature differences are associated with larger gains from local refinement, while the instance-prior gain shows little association.
After adjusting the rank correlations for scene, log observation count, and target-mask size, the coefficients are $0.106$ for instance observation selection and $0.231$ for the additional local-evidence gain.
To account for queries that refer to the same target, we also merge queries sharing instance--scale identities. The resulting $64$ groups retain the qualitative pattern, although these groups do not constitute physical-object ground truth.
The association varies across scenes, and feature distance can reflect changes in visibility and mask content as well as viewpoint-dependent appearance.

\paragraph{Gaussian-local evidence aggregation.}
We test whether the same learned evidence distribution can be summarized by a single feature without reducing segmentation accuracy. The main component analysis compares fixed-mass aggregation with probability-weighted mean and maximum relevance; here we instead average the prototype features before scoring the query.
The mean prototype feature variant in Tab.~\ref{tab:app_aggregation} averages prototypes using their learned probabilities as weights, then normalizes the resulting feature before computing relevance. With the same instance prior and learned distribution at $\beta=0.5$, it yields $67.67$ mIoU, compared with $70.59$ for fixed-mass aggregation.
This result supports using the distribution to select query-relevant prototype support rather than reducing it to a mean feature. The comparison differs from averaging relevance scores or the original Gaussian-associated observation features evaluated in the main table.

\paragraph{Observation reliability and fusion.}
We examine how strongly local evidence should modify the instance prior, and whether it should also be allowed to lower that prior. Observation reliability controls the strength of the correction using observation count and assignment confidence; we compare these two factors under otherwise identical settings.
Removing reliability weighting gives $70.23$ mIoU, compared with $70.59$ for the full rule.
Count-only and assignment-confidence-only weights give $70.41$ and $70.22$, respectively.
The four weighting variants differ by less than $0.4$ mIoU points in this run. The combined weight accounts for both observation count and assignment confidence.

We also compare positive-only fusion with signed correction, $b_g(t)+\rho_g[e_g(t)-b_g(t)]$, keeping observation selection, fixed-mass aggregation, and reliability weights unchanged.
Signed correction also allows local evidence to lower the prior, unlike the positive-only rule in Eq.~\ref{eq:fusion}.
Positive-only fusion yields a slightly higher mean mIoU ($70.59$ vs.\ $70.30$), although the relative performance varies across scenes (Tab.~\ref{tab:app_aggregation}).

\begin{table}[t]
\centering\small
\setlength{\tabcolsep}{3pt}
\caption{\textbf{Aggregation and evidence representation (LERF-OVS mIoU, \%).} Matched single-run comparisons at $\beta=0.5$. Blue rows mark the reference; bold/underline mark the best/second-best Mean within each block (descriptive ranks).}
\label{tab:app_aggregation}
\begin{tabular}{lrrrrr}
\toprule
Variant & Fig. & Tea. & Ram. & Wal. & Mean $\uparrow$ \\
\midrule
\rowcolor{black!6}
\multicolumn{6}{l}{\emph{Local evidence alternatives}} \\
\rowcolor{blue!5}
EviSplat (fixed mass) & 74.53 & 73.53 & 62.83 & 71.45 & \textbf{70.59} \\
Mean prototype feature & 66.61 & 73.10 & 65.40 & 65.56 & \underline{67.67} \\
\midrule
\rowcolor{black!6}
\multicolumn{6}{l}{\emph{Observation reliability and fusion}} \\
\rowcolor{blue!5}
EviSplat (fixed mass) & 74.53 & 73.53 & 62.83 & 71.45 & \textbf{70.59} \\
No weighting ($\rho_g=1$) & 74.57 & 74.60 & 61.74 & 70.02 & 70.23 \\
Count only & 74.31 & 73.49 & 62.06 & 71.77 & \underline{70.41} \\
Assignment confidence only & 74.57 & 74.61 & 62.06 & 69.64 & 70.22 \\
Signed local correction & 74.67 & 73.07 & 61.53 & 71.92 & 70.30 \\
\bottomrule
\end{tabular}
\end{table}

\paragraph{Parameter sensitivity.}
We examine how accuracy changes when the method selects more observations, includes more local probability mass, or assigns Gaussians more broadly across instances.
Each block of Tab.~\ref{tab:app_sensitivity} varies one parameter at a time around $(\beta,\gamma,T)=(0.5,0.25,0.07)$.
The highest mean among the tested observation fractions occurs at $\beta=0.25$ for the instance-only variant ($67.19$) and $\beta=0.5$ for the full model ($70.59$).
At fixed $\beta=0.5$, the full-model mean is highest at $\gamma=0.25$ among the tested mass budgets.
For the soft assignment of Eq.~\plaineqref{eq:assignment}, hard assignment and $T=0.03,0.07$ give similar full-model means of $70.38$, $70.42$, and $70.59$.
The more diffuse assignments at $T=0.15$ and $0.30$ reduce the mean to $68.44$ and $65.73$, with the largest scene-level decreases on \textsc{waldo\_kitchen}.
For the instance-only variant, hard assignment gives a slightly higher mean than $T=0.07$. Together, these results show that the preferred observation fraction depends on whether local evidence is used, and that overly diffuse instance assignments reduce accuracy in this sweep.

\begin{table}[t]
\centering\small
\setlength{\tabcolsep}{5pt}
\caption{\textbf{Parameter sensitivity (LERF-OVS mean mIoU, \%).} Single-run sweeps around $(\beta,\gamma,T)=(0.5,0.25,0.07)$, varying one factor at a time. Blue rows mark the reference setting; bold/underline indicate best/second-best within each parameter block and column.}
\label{tab:app_sensitivity}
\begin{tabular}{llrr}
\toprule
Factor & Value & Instance only $\uparrow$ & Full model $\uparrow$ \\
\midrule
$\beta$ & max observation relevance & 59.99 & 62.35 \\
 & $0.10$ & \underline{66.57} & 68.15 \\
 & $0.25$ & \textbf{67.19} & \underline{69.93} \\
\rowcolor{blue!5}
 & $0.50$ (reference) & 66.22 & \textbf{70.59} \\
 & $1.00$ (mean relevance) & 62.84 & 68.36 \\
\midrule
$\gamma$ & $0.10$ & -- & 68.96 \\
\rowcolor{blue!5}
 & $0.25$ (reference) & -- & \textbf{70.59} \\
 & $0.50$ & -- & \underline{69.63} \\
 & $1.00$ (expectation) & -- & 67.50 \\
\midrule
$T$ & hard assignment & \textbf{66.76} & 70.38 \\
 & $0.03$ & \underline{66.53} & \underline{70.42} \\
\rowcolor{blue!5}
 & $0.07$ (reference) & 66.22 & \textbf{70.59} \\
 & $0.15$ & 59.45 & 68.44 \\
 & $0.30$ & 54.33 & 65.73 \\
\bottomrule
\end{tabular}
\end{table}

\paragraph{Evidence-field training seeds.}
To check whether the observation-fraction comparison is consistent across training seeds, we evaluate $\beta=0.25$ and $\beta=0.5$ on the same three trained fields. Only the observation fraction changes; the fields and remaining inference settings are fixed.
The means are $69.89\pm0.04$ and $70.67\pm0.10$ mIoU, respectively (population standard deviations).
The paired gains are $0.65$, $0.92$, and $0.78$ points.
Across seeds, \textsc{waldo\_kitchen} improves by $3.81$ points and \textsc{teatime} by $0.43$, while \textsc{figurines} and \textsc{ramen} decrease by $0.71$ and $0.40$.
This supports the observed mean improvement across evidence-field seeds, rather than a uniform scene-level gain.
We use $\beta=0.5$ in the main LERF-OVS results following this benchmark sweep; these evaluations do not constitute a held-out test of the parameter choice.

\subsection{Representation Size, Storage, and Computational Cost}

\paragraph{Local representation storage.}
Let $N$ be the number of Gaussians, $L$ the number of semantic scales, and $d_c$ the visual-feature dimension.
The evidence field stores $ND+LDK$ trainable scalars and $LKd_c$ frozen codebook scalars. The latent dimension $D$ controls the per-Gaussian storage budget, and the same latent is decoded at each scale.
In comparison, dense empirical distributions require $NLK$ probability values. A sparse representation stores only supported bins and their indices: at scale $\ell$, their number satisfies $s_g^{(\ell)}\leq\min(K,|\mathcal E_g^{(\ell)}|)$. For fixed geometry and codebooks, field storage is determined by the latent dimension, whereas sparse storage depends on the number of supported prototype bins.
Both representations additionally use the instance observation bank and affinity field.

The storage comparison in Tab.~\ref{tab:lerf_component} measures logical payloads for the local representation: sparse probability values with their CSR indices and offsets, or Gaussian latents with decoder weights.
All rows exclude the common observation bank ($190.24$ MiB), codebooks ($3.00$ MiB), and reliability values ($32.47$ MiB), as well as the affinity field and geometry.
The stored latents and sparse probabilities use FP16; decoders use FP32, and CSR prototype indices and row offsets use uint8 and int32, respectively. These are payload measurements rather than serialized checkpoint sizes or complete system storage.

\paragraph{Evidence-field size and accuracy.}
We compare the learned evidence field with directly stored empirical distributions to test whether useful local evidence can be retained with a smaller storage budget.
Replacing the decoded $p_g$ with the empirical $q_g$, while keeping the prior, reliability, and fixed-mass aggregation unchanged at $\beta=0.5$, gives $70.15$ mIoU versus $70.59$ for the $D=32$ field in the matched single run.
The $D=8$ field uses $86.7$ MiB instead of $307.7$ MiB for sparse empirical distributions, a $71.8\%$ reduction in local payload, with mean mIoU decreasing to $69.29$.
Table~\ref{tab:lerf_component} shows that $D=16$ retains comparable mean mIoU to sparse storage ($70.2$ vs.\ $70.1$) using $173.3$ rather than $307.7$ MiB, a $43.7\%$ reduction. This provides a smaller local representation with similar accuracy in the measured comparison.
The $D=32$ field uses $346.7$ MiB, exceeding sparse storage; its single-run accuracy above is distinct from the three-seed mean reported in the main table.

\begin{figure}[t]
\centering
\includegraphics[width=\linewidth]{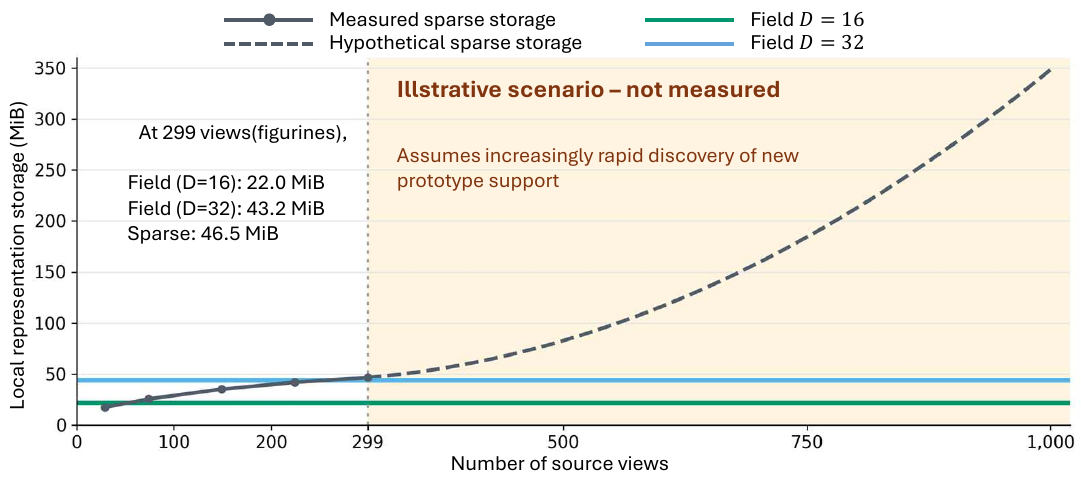}
\caption{\textbf{Local representation storage as observations accumulate.} Solid markers show measurements on \textsc{figurines}. The shaded continuation is hypothetical, not measured or predicted. Field capacity assumes fixed geometry and codebook size; shared storage is excluded.}
\label{fig:app_view_storage}
\end{figure}

\paragraph{Storage with increasing source views.}
A new observation increases sparse storage only if it adds support to a previously unused prototype bin. In contrast, field storage is fixed once the Gaussian set, codebooks, and latent dimension are fixed.
Figure~\ref{fig:app_view_storage} illustrates this distinction on \textsc{figurines}: the measured sparse payload grows from $17.4$ to $46.5$ MiB over $29$--$299$ source views, while the $D=16$ field uses $22.0$ MiB.
The dashed continuation beyond $299$ views assumes increasingly rapid discovery of new prototype support; it is illustrative rather than a measured or predicted trend.
With fixed codebooks, support is bounded by $K$ bins per Gaussian and scale. The observation bank and other shared components are excluded, so fixed field storage does not imply fixed total system storage.

\paragraph{Computational complexity, query time, and training cost.}
We report the computation and training required by the evidence field alongside its storage savings.
Decoding all distributions costs $O(NLDK)$.
The decoded probabilities are query-independent and can be reused across queries, either within processing chunks or in a full $O(NLK)$ cache.
For each query, prototype scoring costs $O(LKd_c)$, sorting costs $O(LK\log K)$, and local aggregation costs $O(NLK)$.
Instance-observation scoring, pooling, soft assignment, and rendering incur additional costs.
Tab.~\ref{tab:app_efficiency} reports measured local query and evidence-field training costs.

The costs in Tab.~\ref{tab:app_efficiency} were measured on a dedicated NVIDIA B200 with no overlapping CUDA workload.
These measurements concern field preparation, training, and the local query computation, which does not depend on the instance observation fraction $\beta$.
Local query time averages each scene's median over its queries and three passes, excluding warmup, text encoding, instance-prior computation, fusion, rendering, and I/O.
The per-query block shows that sparse aggregation takes $6.90$ ms, versus $12.13$ and $13.34$ ms for the $D=8$ and $D=32$ fields, which include fused decoding and local aggregation.
The sparse implementation visits occupied bins, while the field produces $K$ logits per Gaussian; the field is slower in this measured local-query comparison.

The additional $10$k-step field optimization averages $31.21$ s for $D=8$ and $29.16$ s for $D=32$; these measurements do not establish a runtime ordering between the dimensions.
Cached preparation is reported separately and excludes construction of the affinity field and the evidence-set association cache. Optimization timing also excludes checkpoint serialization.
Peak allocated memory includes the resident training caches, with maxima of $5.98$ and $6.56$ GiB across scenes.
Raw empirical distributions need no field optimization; their cache-construction cost is not measured by this timing comparison.

\begin{table}[t]
\centering\small
\setlength{\tabcolsep}{4pt}
\caption{\textbf{Evidence-field computation cost.} Preparation, optimization, and local query times average scenes. Peak memory is the maximum over scenes. Lower is better; timings are not ranked.}
\label{tab:app_efficiency}
\begin{tabular}{lrrr}
\toprule
Quantity $\downarrow$ & Sparse $q_g$ & Field $D=8$ & Field $D=32$ \\
\midrule
\rowcolor{black!6}
\multicolumn{4}{l}{\emph{One-time field preparation and training}} \\
Cached preparation (s) & -- & 17.00 & 14.07 \\
10k-step optimization (s) & -- & 31.21 & 29.16 \\
Peak allocated memory (GiB) & -- & 5.98 & 6.56 \\
\midrule
\rowcolor{black!6}
\multicolumn{4}{l}{\emph{Per-query local computation}} \\
Local query time (ms) & 6.90 & 12.13 & 13.34 \\
\bottomrule
\end{tabular}
\end{table}

\clearpage
\section{Qualitative Results}
\label{app:qualitative}

\begin{figure}[htbp]
\centering
\includegraphics[width=\linewidth]{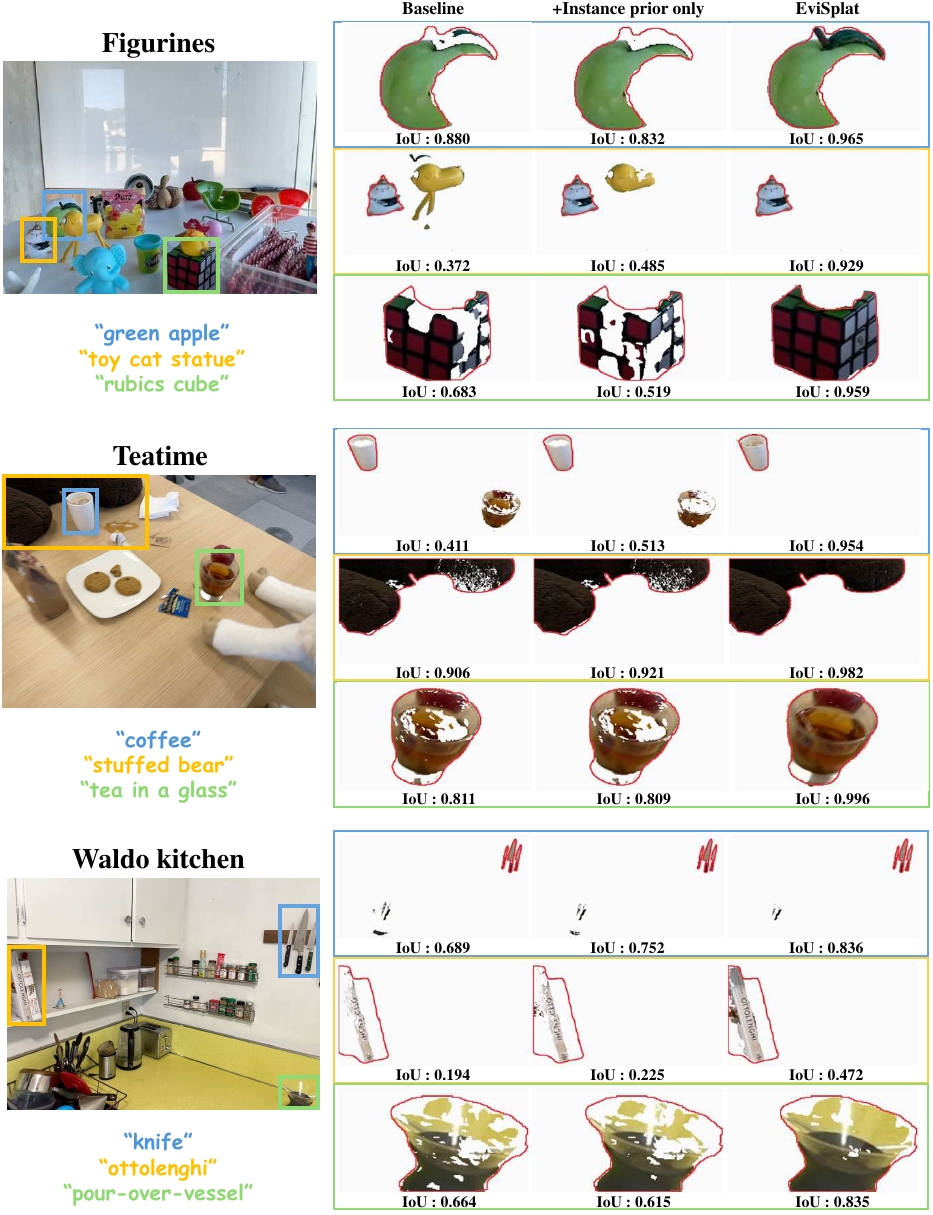}
\caption{\textbf{Qualitative results on LERF-OVS.} Baseline, instance prior only, and EviSplat. Red outlines mark ground truth; values report rendered-view mask IoU.}
\label{fig:app_qualitative}
\end{figure}

\begin{figure}[htbp]
\centering
\includegraphics[width=\linewidth]{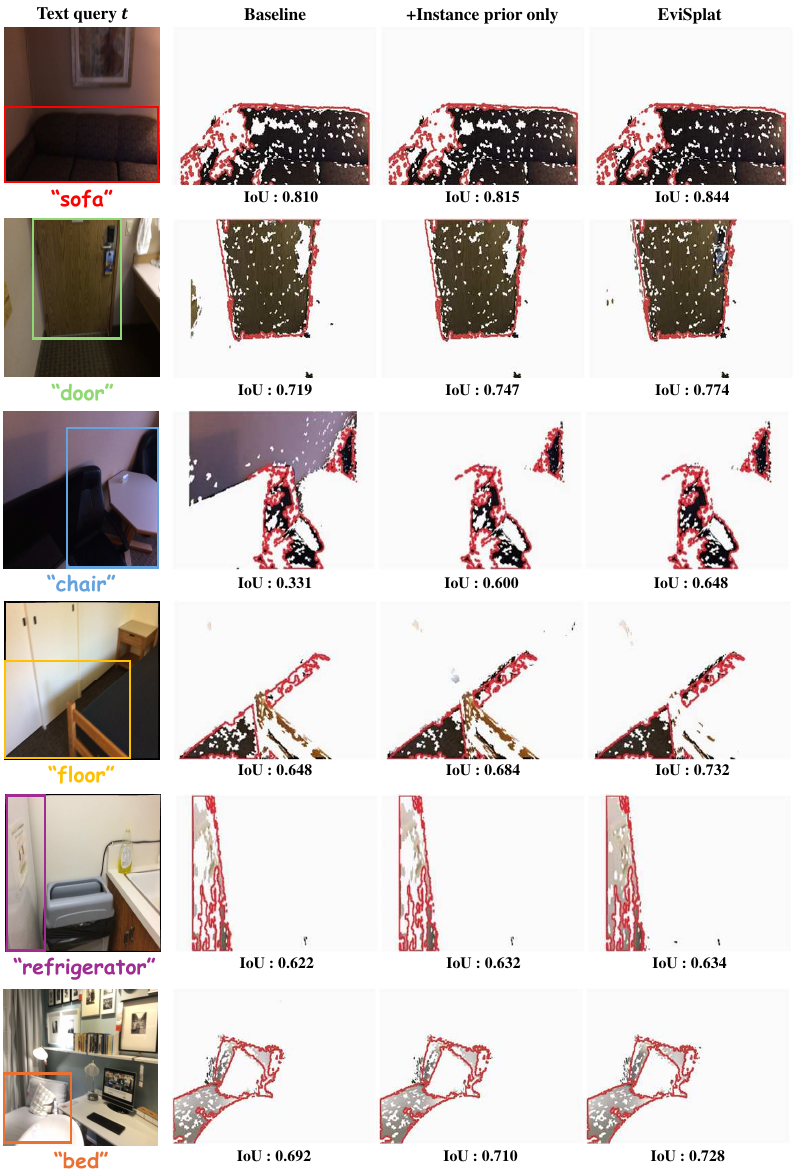}
\caption{\textbf{Qualitative results on ScanNet.} Baseline, instance prior only, and EviSplat, shown from selected views. Values report query-specific point IoU over the full scene.}
\label{fig:app_scannet_qualitative}
\end{figure}

Figures~\ref{fig:app_qualitative} and~\ref{fig:app_scannet_qualitative} compare the baseline, instance prior only, and EviSplat on LERF-OVS and ScanNet, respectively.
The examples show changes in target coverage and false-positive regions as instance observations and Gaussian-local evidence are introduced.
The gains are not uniform: for ``rubics cube'' and ``pour-over vessel'' on LERF-OVS, the instance prior alone falls below the baseline, while adding Gaussian-local evidence recovers the target.

\end{document}